\documentclass[11pt,a4paper]{article}
\usepackage{aes} %
\usepackage{amsmath,amssymb}
\usepackage{array,tabularx}
\usepackage{tikz}\usetikzlibrary{arrows.meta,positioning,calc,backgrounds,shadows.blur,shapes.callouts,decorations.pathreplacing}
\usepackage{fontawesome5}
\tcbuselibrary{skins,breakable}
\input{appendix/traces/tracestyle}
\newfontfamily\trplexsans{IBMPlexSans}[Extension=.otf, UprightFont=*-Regular, BoldFont=*-Bold,
  ItalicFont=*-Italic, BoldItalicFont=*-BoldItalic, FontFace={mb}{n}{*-Medium}]
\newfontfamily\trplexmono{IBMPlexMono}[Extension=.otf, UprightFont=*-Regular, BoldFont=*-Bold,
  ItalicFont=*-Italic, BoldItalicFont=*-BoldItalic, FontFace={mb}{n}{*-Medium}]
\renewcommand{\trfontsans}{\trplexsans}
\renewcommand{\trfontmono}{\trplexmono}
\renewcommand{\trmedium}{\fontseries{mb}\selectfont}
\newcommand{\conceptsizes}{\renewcommand{\footnotesize}{\fontsize{8}{9.5}\selectfont}%
  \renewcommand{\scriptsize}{\fontsize{7}{8}\selectfont}\renewcommand{\tiny}{\fontsize{5}{6}\selectfont}}
\newcommand{\paragraphtight}[1]{\textbf{#1~}}

\usepackage{caption}
\title[Metrics in the Loop]{Metrics-in-the-Loop: Complexity Feedback Reshapes Agent-Written Code}

\aesauthor[affil={1,3}]{Nicolas Chapados}
\aesauthor[affil={1}]{Jean-François Gagné}
\aesauthor[affil={2,3,4}]{Nouha Dziri}
\aesaffiliation{1}{Nera Software}
\aesaffiliation{2}{Cohere Labs}
\aesaffiliation{3}{Mila\;--\;Quebec Artificial Intelligence Institute}
\aesaffiliation{4}{ELLIS Institute Tübingen}
\aesemails{\aesemail{chapados@nera.software}\quad \aesemail{jfgagne@nera.software}\quad \aesemail{nouha.dziri@cohere.com}}

\begin{document}
\maketitle
\suppressfloats[t] %

\begin{abstract}
Coding agents now write whole codebases in long autonomous runs, guided mainly by whether their tests pass.
Yet passing tests alone does not reveal accumulating complexity or gaps in test coverage, which then persist as agents extend their own code.
We introduce metrics-in-the-loop, an inference-time intervention that gives agents in-flight feedback: cheap, automatically computed code metrics injected into the agent's context during the trajectory. After each code-changing agent turn, we insert short feedback messages naming changed functions whose cyclomatic complexity exceeds a limit (complexity feedback) or changed lines that are skipped by tests (coverage feedback), giving the agent the opportunity to refactor code or strengthen tests before continuing. We run extensive experiments and ablations comparing several interventions across three long-horizon coding benchmarks and four strong models. We find that complexity feedback significantly reduces erosion, a measure of complexity concentration of code, without affecting held-out correctness of the code. Broadly, our experiments investigate a practical form of agent ``re-railing'', where timely, actionable measurements redirect agents before weaknesses become the foundation they keep building on. As agents scale to codebases with millions of lines, such interventions become critical to sustaining effective development and long-term code maintainability.

\end{abstract}

\section{Introduction}
\label{sec:introduction}
\begin{figure}[t]
\centering
\begin{minipage}[c]{0.44\linewidth}\centering
{\conceptsizes\resizebox{\linewidth}{!}{\begin{tikzpicture}[
  font=\trfontsans\footnotesize, >={Stealth[length=2.1mm, width=1.7mm]},
  card/.style={rounded corners=1.6mm, fill=#1Bg, draw=#1, line width=0.5pt},
  bar/.style={rounded corners=0.5mm, inner sep=0pt},
  flow/.style={->, line width=0.9pt, draw=#1!85, rounded corners=2.5mm},
  small/.style={font=\trfontsans\tiny, text=trMuted}]

  \node[card=trAgent, minimum width=56mm, minimum height=19mm, anchor=north west] (ed) at (0,0) {};
  \draw[trAgent!30, line width=0.4pt] ([yshift=-5mm]ed.north west) -- ([yshift=-5mm]ed.north east);
  \node[anchor=west, text=trAgent!90!black, font=\trfontsans\trmedium\scriptsize] at ([xshift=2mm, yshift=-2.5mm]ed.north west)
    {\faRobot\ \ coding agent, turn $t$};
  \foreach \k in {0,1,2} \fill[trAgent!30] ([xshift=-3mm-\k*2.4mm, yshift=-2.5mm]ed.north east) circle (0.7mm);
  \fill[trFeedbackBg, rounded corners=1mm] (2mm,-10.2mm) rectangle (54mm,-17.6mm);
  \draw[trFeedback!70, line width=0.4pt, rounded corners=1mm] (2mm,-10.2mm) rectangle (54mm,-17.6mm);
  \node[anchor=south west, font=\trfontmono\tiny, text=white, fill=trFeedback, rounded corners=0.8mm, inner sep=1.5pt]
    at (3mm,-9.8mm) {def visit\_Template(...)};
  \node[anchor=south east, small, text=trFeedback!80!black] at (54mm,-9.8mm) {116 lines, CC 20};
  \foreach \y/\w/\x in {-7/22/4, -12/30/7, -14/40/10, -16/24/13} {
    \fill[trBefore!45, bar] (\x mm,\y mm) rectangle ++(\w mm, 0.9mm);
  }

  \node[card=trKey, minimum width=56mm, minimum height=29mm, below=8mm of ed] (sk) {};
  \draw[trKey!30, line width=0.4pt] ([yshift=-5mm]sk.north west) -- ([yshift=-5mm]sk.north east);
  \node[anchor=west, text=trInk, font=\trfontsans\trmedium\scriptsize] at ([xshift=2mm, yshift=-2.5mm]sk.north west)
    {\faTachometer*\ \ SlapKit measures the change};
  \draw[flow=trAgent] (ed.south) -- node[right, small, text=trAgent!70!black, align=left]
    {turn ends:\\\faCodeBranch\ changes this turn} (sk.north);
  \begin{scope}[shift={([xshift=6mm, yshift=-21mm]sk.north west)}]
    \node[anchor=south west, font=\trfontsans\trmedium\tiny, text=trInk] at (-1mm,12.5mm) {complexity (CC)};
    \foreach \i/\h/\c in {0/20/trFeedback, 1/4/trAgent, 2/3/trAgent, 3/2/trAgent}
      \fill[\c!85, rounded corners=0.5mm] (\i*4.4mm,0) rectangle ++(3mm, \h*0.55mm);
    \node[font=\trfontsans\trmedium\tiny, text=trFeedback!85!black, right] at (3mm,10mm) {20};
    \draw[dashed, trFeedback!70!black, line width=0.6pt] (-1mm,3.3mm) -- (17.5mm,3.3mm);
    \node[font=\trfontsans\tiny, text=trFeedback!70!black, above left] at (17.5mm,3.3mm) {limit 6};
    \draw[trBefore!60, line width=0.4pt] (-1mm,0) -- (17.5mm,0);
    \node[small, below] at (8mm,0) {changed functions};
  \end{scope}
  \begin{scope}[shift={([xshift=36mm, yshift=-21mm]sk.north west)}]
    \node[anchor=south west, font=\trfontsans\trmedium\tiny, text=trInk] at (-2mm,12.5mm) {coverage};
    \foreach \r in {0,...,2} \foreach \c in {0,...,2} {
      \pgfmathtruncatemacro{\n}{\r*3+\c}
      \ifnum\n<6 \fill[trAfter!80] (\c*4mm, 9.6mm-\r*3.6mm) circle (1.1mm);
      \else \draw[trFeedback, line width=0.5pt, fill=white] (\c*4mm, 9.6mm-\r*3.6mm) circle (1.05mm); \fi
    }
    \node[small, below] at (4mm,0) {changed lines};
  \end{scope}
  \node[font=\trfontsans\tiny, text=trMuted, anchor=south] at ([yshift=0.8mm]sk.south)
    {\faDatabase\ \ unchanged files: complexity from cache};

  \node[rectangle callout, callout absolute pointer={([yshift=-0.4mm]sk.south)}, rounded corners=2.4mm,
        fill=trFeedbackBg, draw=trFeedback, line width=0.5pt, align=left, font=\trfontmono\tiny, inner sep=3.5pt,
        text width=50mm,
        below=5.5mm of sk] (msg)
    {\textsf{\color{trFeedback!85!black}\faCommentDots}\ \ [Measurement, turn 92]\\
     Functions above the limit, highest first:\\
     \ compiler.py:761\ \ CodeGenerator.visit\_Template\\
     \ \ cyclomatic complexity 20 (limit 6)};

  \draw[flow=trFeedback] (msg.west) -- ++(-6mm,0) |- node[pos=0.25, sloped, above, font=\trfontsans\scriptsize,
        text=trFeedback!85!black] {added before turn $t+1$} ([yshift=-7mm]ed.west);
\end{tikzpicture}}}\\[1pt]{\small (a)}
\end{minipage}\hfill
\begin{minipage}[c]{0.52\linewidth}\centering
\includegraphics[width=\linewidth]{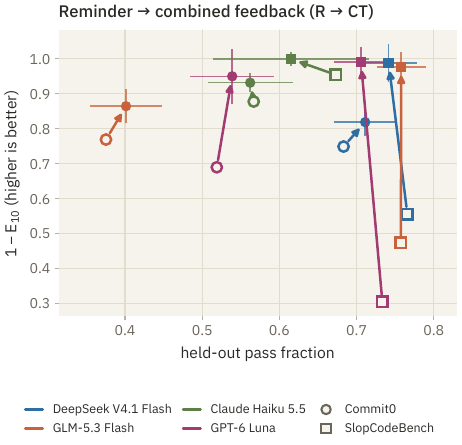}\\[1pt]{\small (b)}
\end{minipage}
\caption{Metrics-in-the-loop. \textbf{(a)}~One round: after a turn that changes code, the changed functions are measured and the report (cyclomatic complexity (C) and/or test coverage (T)) enters the agent's context before its next turn. \textbf{(b)}~Adding both reports to the reminder baseline (R$\to$CT) lowers erosion for all four models on Commit0 (circles) and SlopCodeBench (squares), with no detectable effect on held-out correctness for any model. Each arrow starts at R (hollow) and shows the estimated paired change to CT (filled), with whiskers for the 95\% intervals.
The axes show held-out pass fraction and one minus the share of complexity in functions above 10 (higher is better for both axes).}%
\label{fig:mitl-loop}
\end{figure}
Software engineering involves more than writing code that works: code must also be organized so that it remains testable and easy to change as requirements evolve~\citep{parnas1972criteria,lehman1980programs}. Coding agents now write code faster than any human can review, and how agents write often departs from well-engineered human code, favoring large, monolithic, single-file implementations~\citep{yang2026programbench}. The problem compounds when an agent builds on its own work: successive extensions concentrate complexity and duplicate code~\citep{orlanski2026slop}, and evolving software across many files without breaking what already works remains difficult~\citep{le2025swe,deng2026swemilestone}. As agent-written codebases grow to millions of lines, rising complexity invites bugs and vulnerabilities, making maintenance progressively harder~\citep{banker1993complexity,parnas1994aging,eick2001decay}.

We ask whether cheap, mechanical feedback delivered during code generation can improve code
structure without sacrificing functional correctness. We introduce MITL (Metrics-In-The-Loop), a
lightweight plugin that measures the code an agent has just changed, provides feedback to the agent in the form of a short report before the agent continues, and requires no training, no reference solution and no access to held-out tests (Figure~\ref{fig:mitl-loop}a).
The report covers: (1) cyclomatic complexity~\citep{mccabe1976complexity}, the number of independent paths through a function, flagging every function the agent has just modified whose complexity exceeds a threshold (6 in our experiments); and (2) test coverage, counting for each modified function the lines that the agent's own tests do not execute~\citep{inozemtseva2014coverage}.

The report alone, delivered often, could also change the agent's behavior. We therefore compare complexity feedback, coverage feedback and both combined against a reminder-only control and a no-message baseline (Section~\ref{sec:design}). We estimate effects separately for each of four models of varying capability, open-weight and proprietary (DeepSeek V4.1 Flash, GLM-5.3 Flash, Claude Haiku 5.5 and GPT-6 Luna), on three long-horizon coding benchmarks (Commit0~\citep{zhao2025commit0}, SlopCodeBench~\citep{orlanski2026slop} and SWE-EVO~\citep{le2025swe}) and also report pooled results.

Our results show that complexity feedback, alone or combined with coverage feedback, lowers erosion, the share of complexity concentrated in functions with complexity above 10. The reduction is significant for all four models on every benchmark except SWE-EVO, where it is smaller and significant for two of the four models. We also carefully audit agent traces to understand their behavior: agents act on the feedback, reworking most of the functions named therein. In short, cheap, actionable measurements lead agents to write structurally simpler code without damaging the correctness of the code (Table~\ref{tab:main}, Figure~\ref{fig:mitl-loop}b).

As coding agents write a growing share of the world's software, it is pressing to evaluate not only whether their code works but what they leave behind. Our work takes a realistic look at in-run feedback: cheap metrics steer agents toward simpler structure. Ultimately, we expect better code quality to translate into better correctness, but current benchmarks cannot yet demonstrate this~\citep{trivedi2026cleanliness}, underscoring the importance of benchmark design and work ahead for the community. Building agents that sustain codebases of millions of lines will require evaluations that measure maintainability as rigorously as correctness. %

\section{Experimental design}
\label{sec:design}
\paragraphtight{The loop.} The coding agent receives a task and is instructed to complete it in a sandbox, through Oh-My-Pi~\citep{boluk2026ohmypi}, the harness that provides the model's tools and manages its memory. After each turn that changes a file, the harness saves a workspace snapshot and SlapKit, our measurement tool, measures the functions the turn changed and emits a report, which enters the agent's context as feedback before its next model call. Complexity is measured as cyclomatic complexity~\citep{mccabe1976complexity}, which counts the independent paths through a function: the more paths, the harder the function is to follow and to test~\citep{curtis1979measuring}.
The \emph{complexity report} lists changed functions with cyclomatic complexity above 6, in descending order, and asks the agent to bring them under this limit. For coverage, SlapKit reruns the agent's own tests; the \emph{coverage report} lists changed functions with added or edited lines that these tests do not execute, with a target of zero uncovered changed lines; the agent is then free to write or modify tests to bring down the count of uncovered lines. %

\paragraphtight{Arms.} We introduce five experimental arms to distinguish the effects of measurements and feedback instructions. B0 receives no feedback. R receives the feedback instructions alone with no measurements (a ``placebo feedback'' of sorts), and C, T and CT respectively add the complexity,  coverage, or both reports (see Appendix~\ref{app:design} for details). R, C, T and CT share byte-identical feedback instructions and the same delivery rule: each receives one message after every turn that changes code and can be measured. Example feedback is reported in Appendix~\ref{app:messages}. We compute the following contrasts: CT$-$R estimates the effect of adding both reports to the reminder, C$-$R and T$-$R isolate each, and R$-$B0 measures the reminder alone. %
The estimated effects also include changes to the message header and a task-prompt paragraph defining each metric. Report counts may also vary by arm because feedback can affect how often the code can be measured within the common budget.

\paragraphtight{Models and benchmarks.} We evaluate four models of varying capability, all at high reasoning effort: the open-weight DeepSeek V4.1 Flash (released 2026-09-10) and GLM-5.3 Flash (released 2026-08-26), and the proprietary Claude Haiku 5.5 and GPT-6 Luna.
Each model runs every arm on three long-horizon Python benchmarks. Commit0~\citep{zhao2025commit0} asks agents to reconstruct Python libraries from specifications; we use 25 libraries, three to five seeds per arm and a 2M processed-token budget per run. SlopCodeBench~\citep{orlanski2026slop} evaluates program development over successive specifications; we use 35 problems with three to eight checkpoints, two seeds and 1M tokens per checkpoint. Each checkpoint extends the specification and resets the context, retaining the workspace. SWE-EVO~\citep{le2025swe} asks agents to update published projects to their next release; we use 35 of 48 transitions, two seeds, and a budget of 2M tokens and 90 minutes per run.

\paragraphtight{Outcomes.}\label{sec:outcomes} We measure held-out correctness by the fraction of a fixed test set that the final code passes. The agent never sees these tests: they are stored outside the repository the agent is permitted to access. Commit0's official suite grades the last snapshot within the token cap, while the agent writes its own tests during the run; SlopCodeBench grades the final checkpoint with hidden tests; SWE-EVO grades the final code by its fix rate on the next release's withheld tests: the fraction of tests failing on the starting code that pass on the final code, set to zero if the agent breaks any test that passes both before and after the reference change. For structural complexity, we adapt SlopCodeBench's \emph{erosion index} $E_{\tau}$: with $c_f$ the cyclomatic complexity of function $f$ and $n_f$ its count of nonblank, noncomment lines,
\begin{equation}
E_{\tau} = \frac{\sum_f c_f\sqrt{n_f}\,\mathbf{1}\{c_f>\tau\}}{\sum_f c_f\sqrt{n_f}}
\label{eq:erosion}
\end{equation}
is the share of complexity mass in functions with complexity above $\tau$. The summation only include the main codebase functions in this computation, excluding test-suite members. For Commit0 and SlopCodeBench, we report $E_{10}$. For SWE-EVO, since the agent edit an existing project, whose original code still makes up most of the final tree, we report the change in $E_{10}$ from the starting code. These changes are small, so we give them in thousandths (\textperthousand); for scale, the starting code's $E_{10}$ averages 330\textperthousand.  
Appendix~\ref{app:design} gives cell counts, exclusions and the integrity checks.

\section{Results}
\label{sec:results}
\begin{figure}[t]
\centering
\includegraphics[width=0.9\linewidth]{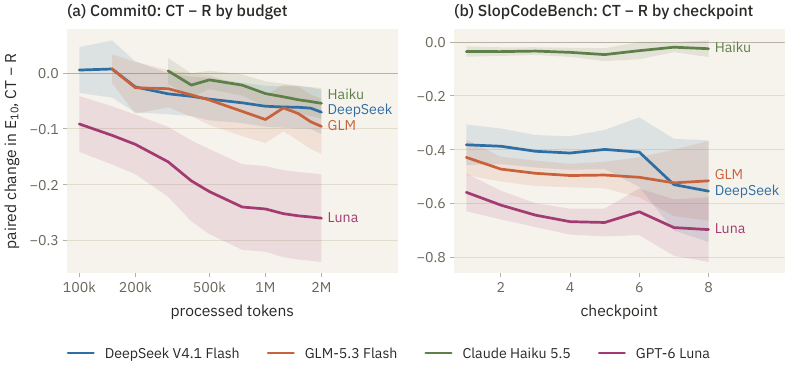}
\caption{CT$-$R in $E_{10}$ during the run, by model, with pointwise 95\% $t$ bands. \textbf{(a)}~Commit0: paired differences within libraries at each token budget (log scale, up to 2M), with at least 15 paired libraries. \textbf{(b)}~SlopCodeBench: paired differences within problems at each checkpoint, with at least 5 pairs; not all problems reach 8 checkpoints.}
\label{fig:trajectories}
\end{figure}

\paragraphtight{Complexity feedback lowers erosion.} C and CT significantly lower final $E_{10}$ relative to R for all four models on Commit0 and SlopCodeBench, with every interval excluding zero (Table~\ref{tab:main}). On SlopCodeBench, both arms keep erosion at or below 0.022 at every checkpoint. Effects vary by model partly reflecting differences in reminder-arm erosion but all pooled intervals exclude zero with a reduction range of 7.8--24.6 percentage points of starting erosion. On SWE-EVO, all models have lower estimated erosion, but only two models' intervals exclude zero for each contrast under project clustering. 
Appendix~\ref{app:tables} reports results for all contrasts.
The CT$-$R erosion gap grows during Commit0 runs and persists across SlopCodeBench's context resets (Figure~\ref{fig:trajectories}; Appendix~\ref{app:trajectories}). These results show that complexity feedback can steer agents toward simpler code structures as they extend their implementations over long coding runs.

\begin{table}[t]
\centering\footnotesize
\setlength{\tabcolsep}{3pt}
\caption{Erosion under complexity feedback and correctness under combined feedback, by model. $E_{10}$ ($\downarrow$~better) is the share of complexity mass in functions with cyclomatic complexity above 10. Contrasts are paired mean differences with 95\% intervals. Correctness ($\uparrow$~better) is the held-out pass fraction or, on SWE-EVO, the fix rate; differences are in percentage points. Commit0 erosion estimates use seeds 1--5 for R and CT and 1--3 for C. Seeds 4 and 5 were added after the initial analysis; correctness uses the planned seeds 1--3. SWE-EVO intervals are clustered by six projects, with CT using 29 of 35 transitions. Pooled estimates are post hoc. $I^2$~\citep{higgins2002quantifying} estimates the proportion of variation among models attributable to heterogeneity; high $I^2$ indicates that they average effects that differ across models. App.~\ref{app:tables} reports all contrasts and paired samples.}
\label{tab:main}
\begin{tabular}{@{}lrccc@{}}\toprule
 & $E_{10}$ & \multicolumn{2}{c}{change in $E_{10}$} & correctness \\
Model & under R & C--R & CT--R & CT--R \\
\midrule
\multicolumn{5}{@{}l}{\textit{Commit0, 25 libraries}} \\
DeepSeek & 0.251 & \textminus{}0.123 {\scriptsize[\textminus{}0.185, \textminus{}0.061]} & \textminus{}0.070 {\scriptsize[\textminus{}0.110, \textminus{}0.030]} & +4.4 {\scriptsize[\textminus{}1.2, +10.0]} \\
GLM & 0.231 & \textminus{}0.120 {\scriptsize[\textminus{}0.177, \textminus{}0.064]} & \textminus{}0.096 {\scriptsize[\textminus{}0.145, \textminus{}0.046]} & +4.7 {\scriptsize[\textminus{}1.6, +11.0]} \\
Haiku & 0.122 & \textminus{}0.070 {\scriptsize[\textminus{}0.108, \textminus{}0.031]} & \textminus{}0.054 {\scriptsize[\textminus{}0.082, \textminus{}0.026]} & \textminus{}2.3 {\scriptsize[\textminus{}10.2, +5.6]} \\
Luna & 0.311 & \textminus{}0.266 {\scriptsize[\textminus{}0.346, \textminus{}0.186]} & \textminus{}0.261 {\scriptsize[\textminus{}0.339, \textminus{}0.182]} & +3.3 {\scriptsize[\textminus{}4.8, +11.5]} \\
Pool (post hoc) & & \textminus{}0.112 {\scriptsize[\textminus{}0.138, \textminus{}0.087]} & \textminus{}0.078 {\scriptsize[\textminus{}0.097, \textminus{}0.059]} & +3.1 {\scriptsize[\textminus{}0.1, +6.3]} \\
\quad $I^2$ & & 86\% & 89\% & 0\% \\
\addlinespace[5pt] \multicolumn{5}{@{}l}{\textit{SlopCodeBench, 35 problems}} \\
DeepSeek & 0.430 & \textminus{}0.428 {\scriptsize[\textminus{}0.483, \textminus{}0.372]} & \textminus{}0.433 {\scriptsize[\textminus{}0.487, \textminus{}0.379]} & \textminus{}2.4 {\scriptsize[\textminus{}6.1, +1.2]} \\
GLM & 0.530 & \textminus{}0.529 {\scriptsize[\textminus{}0.573, \textminus{}0.485]} & \textminus{}0.504 {\scriptsize[\textminus{}0.546, \textminus{}0.461]} & 0.0 {\scriptsize[\textminus{}3.1, +3.2]} \\
Haiku & 0.046 & \textminus{}0.045 {\scriptsize[\textminus{}0.066, \textminus{}0.025]} & \textminus{}0.045 {\scriptsize[\textminus{}0.065, \textminus{}0.025]} & \textminus{}5.8 {\scriptsize[\textminus{}15.9, +4.3]} \\
Luna & 0.696 & \textminus{}0.692 {\scriptsize[\textminus{}0.733, \textminus{}0.652]} & \textminus{}0.687 {\scriptsize[\textminus{}0.729, \textminus{}0.645]} & \textminus{}2.8 {\scriptsize[\textminus{}6.3, +0.8]} \\
Pool (post hoc) & & \textminus{}0.246 {\scriptsize[\textminus{}0.261, \textminus{}0.230]} & \textminus{}0.236 {\scriptsize[\textminus{}0.252, \textminus{}0.221]} & \textminus{}1.7 {\scriptsize[\textminus{}3.6, +0.1]} \\
\quad $I^2$ & & 100\% & 100\% & 0\% \\
\addlinespace[5pt] \multicolumn{5}{@{}l}{\textit{SWE-EVO, six projects; $E_{10}$ in \textperthousand, and the R column is its change from the starting code}} \\
DeepSeek & +0.15 & \textminus{}2.87 {\scriptsize[\textminus{}5.22, \textminus{}0.51]} & \textminus{}2.30 {\scriptsize[\textminus{}4.96, +0.36]} & +5.1 {\scriptsize[\textminus{}2.3, +12.5]} \\
GLM & +0.21 & \textminus{}11.46 {\scriptsize[\textminus{}31.04, +8.12]} & \textminus{}4.72 {\scriptsize[\textminus{}7.02, \textminus{}2.43]} & +1.5 {\scriptsize[\textminus{}31.2, +34.1]} \\
Haiku & \textminus{}0.10 & \textminus{}5.85 {\scriptsize[\textminus{}14.75, +3.05]} & \textminus{}2.78 {\scriptsize[\textminus{}6.91, +1.34]} & +0.4 {\scriptsize[\textminus{}8.1, +9.0]} \\
Luna & +0.85 & \textminus{}4.93 {\scriptsize[\textminus{}8.46, \textminus{}1.40]} & \textminus{}5.27 {\scriptsize[\textminus{}8.76, \textminus{}1.79]} & +2.2 {\scriptsize[\textminus{}5.4, +9.7]} \\
Pool (post hoc) & & \textminus{}3.68 {\scriptsize[\textminus{}4.95, \textminus{}2.41]} & \textminus{}3.85 {\scriptsize[\textminus{}4.78, \textminus{}2.92]} & +2.7 {\scriptsize[\textminus{}0.2, +5.6]} \\
\quad $I^2$ & & 62\% & 67\% & 0\% \\
\bottomrule\end{tabular}
\end{table}

\paragraphtight{No detectable effect on held-out correctness for any model.} Every per-model correctness interval in Table~\ref{tab:main} includes zero, and on Commit0, CT$-$R runs from $-2.3$ points for Haiku to $+4.7$ for GLM. %
The post hoc pool, $+3.1$ [$-0.1$, $+6.3$], includes zero as well. %
On SlopCodeBench, CT$-$R runs from $-5.8$ points for Haiku to $0.0$ for GLM. The pooled estimate is negative ($-1.7$ [$-3.6$, $+0.1$]), with an interval that also covers zero. %
As a joint mixed model gives an odds ratio of 0.82 [0.70, 0.96], a small cost of the combined arm on SlopCodeBench cannot be ruled out. On SWE-EVO the pooled fix-rate estimate is $+2.7$ [$-0.2$, $+5.6$]. Although modal results suggest a slight average increase in correctness under CT feedback, each benchmark has per-model intervals compatible with losses of 6 percentage points or more. We hypothesize two reasons why simpler code does not yet translate into higher correctness. First, the payoff of lower complexity should grow with scale: as agent-written codebases reach millions of lines, complexity makes them harder to change safely. Even SlopCodeBench, whose agents extend their own code over several checkpoints, involves codebases orders of magnitude smaller, so the effect may not yet show. As agents take on larger codebases, benchmarks must grow with them to reveal how much correctness depends on maintainable code. Second, many failures, such as misread specifications, logic errors or missed edge cases, arise in code that is already simple and passes the agent's own tests, so lowering complexity cannot fix them.

\paragraphtight{Agents act on the reports.} Under C, agents re-edit 54--88\% of flagged functions on Commit0 and 87--94\% on SlopCodeBench. By the end, 46--94\% and 94--97\%, respectively, meet the complexity limit, compared with 1--18\% of over-limit functions edited under R. Agents see at least 98.2\% of offered messages. Coverage feedback also prompts changes, most consistently on SlopCodeBench (Table~\ref{tab:app-response}). The reports thus prompt targeted revisions, with consistently stronger responses to complexity feedback alone than to combined feedback across all models on both benchmarks.

\paragraphtight{Structural changes behind the erosion reduction.} Splitting functions can lower $E_{10}$ without simplifying control flow. After inlining every eligible single-use function in both arms, including library-authored helpers, 3--46\% of the erosion difference remains on Commit0 and 8--18\% on SlopCodeBench for DeepSeek, GLM and Luna (Table~\ref{tab:app-decomposition}). This sensitivity does not isolate extraction's causal effect. On Commit0, total branching changes by less than 4\% for DeepSeek and GLM and falls by 16\% for Luna; on SlopCodeBench, it falls by 14--21\% for all three. Other withheld measures agree on SlopCodeBench: benchmark erosion falls for all models, while verbosity and hidden-suite coverage improve for these three models (Luna's coverage under C only). On Commit0, C has no detectable effect on hidden-suite coverage.

\paragraphtight{The reminder and coverage feedback on their own.} The reminder alone lowers $E_{10}$ on SlopCodeBench by $0.133$ to $0.383$ across models. For Haiku, the reduction from the reminder ($-0.383$ for R$-$B0) is much larger than the additional reduction from complexity feedback ($-0.045$ for C$-$R). On Commit0 it lowers $E_{10}$ for Haiku and Luna only, though DeepSeek's B0 cells there came from another date and another mix of providers. Coverage feedback alone does not lower erosion, and on SlopCodeBench it raises $E_{10}$ for DeepSeek, GLM and Luna (pool $+0.057$ [$+0.044$, $+0.070$], $I^2$ 91\%). Every per-model T$-$R and R$-$B0 correctness interval includes zero, on every benchmark. Two other analyses have intervals excluding zero: pooled SWE-EVO resolution R$-$B0 of $-3.2$ points [$-6.1$, $-0.2$] and, in the mixed model, DeepSeek's R$-$B0 odds ratio on Commit0 (2.26 [1.25, 4.08]).

\section{Limitations and outlook}
\label{sec:conclusion}
Our work shows that MITL feedback, delivered while the agent is writing, can steer agents toward simpler code without sacrificing correctness, laying the groundwork for code that both humans and agents can better understand and extend as codebases grow to millions of lines. Although our experiment does not show a clear gain in correctness, we expect simpler code to remain more correct over time, as it should accumulate fewer errors when an agent keeps extending its own work. Two properties of the current benchmarks likely hide this effect: pass rates are near saturation on SlopCodeBench and dominated by between-library variance on Commit0, with neither benchmark extending the code long enough for structure to pay off, and both are public and therefore subject to contamination. Testing this expectation calls for a new benchmark of unseen specifications with long sequences of change requests, measuring change correctness and cost for code written with and without feedback.

Beyond correctness, our study covers Python, one harness, four models and one budget per benchmark; we plan to broaden each. Reports combine limits and measurements between turns, leaving target, content and timing effects confounded. Evaluating such feedback requires distinguishing effects beyond reminders, structural gains beyond function splitting, and benefits to later work. We address the first two (Table~\ref{tab:lessons}); the proposed benchmark would test the third: which code properties predict subsequent agent failures. Our MITL loop could also carry other measurements, including semantic complexity beyond control flow, adapt its limit to the model, and learn when best to deliver feedback. We inherited our notions of clean code from people, and we do not know whether the same notions should apply to readers with context windows, but as coding agents are entrusted with writing and maintaining a growing fraction of all software, this has become an experimental question worth answering.

\begin{table}[t]
\centering\footnotesize
\setlength{\tabcolsep}{4pt}
\caption{List of checks of interventions made during a coding agent run (Section~\ref{sec:results}, Appendix~\ref{app:design}).}
\label{tab:lessons}
\begin{tabularx}{\linewidth}{@{}>{\raggedright\arraybackslash}p{0.19\linewidth}>{\raggedright\arraybackslash}p{0.21\linewidth}>{\raggedright\arraybackslash}X@{}}
\toprule
Check & Threat it guards against & What it changed here \\
\midrule
Match instructions and delivery in the reminder arm & Instructions or delivery frequency affect behavior & For Haiku on SlopCodeBench, the reminder alone reduces $E_{10}$ by $0.383$; complexity feedback reduces it by a further $0.045$ \\
\addlinespace[2pt]
Outcomes withheld from the agent & Evaluation rewards the optimized metric \citep{manheim2018goodhart} & On SlopCodeBench, the benchmark's erosion score, verbosity and hidden-suite coverage mostly support the $E_{10}$ result; on Commit0, every model's C$-$R coverage interval includes zero \\
\addlinespace[2pt]
Audit the changes behind the score & Function splitting lowers $E_{10}$ without simplifying code & On Commit0, 3 to 46\% of each erosion difference remains after inlining in both arms; DeepSeek's and GLM's total branching changes little \\
\addlinespace[2pt]
Specify the scale; report models before pooling & A pooled estimate obscures model differences & Absolute $E_{10}$ effects on SlopCodeBench differ fifteenfold ($I^2$ 100\%), yet every model removes nearly all erosion present under the reminder arm \\
\addlinespace[2pt]
Identify independent units & Ignored task corre- lation gives too-narrow intervals~\citep{miller2024errorbars} & For SWE-EVO, transition-level intervals exclude zero for seven of eight model estimates across C$-$R (35 transitions) and CT$-$R (29); project clustering reduces this to four \\
\addlinespace[2pt]
Use operating-system isolation for held-out material & Agents access hidden tests or finished code & On Commit0, agents routinely bypassed a filter on path words, before proper exclusion measures were taken;
on SlopCodeBench and SWE-EVO, operating-system isolation (bubblewrap in Dockers) blocked every attempt \\
\addlinespace[2pt]
Assess missing outcomes & Complete-case analysis selects for runs each arm finished & Filling missing SlopCodeBench outcomes with zero or an adverse value changes the pooled CT$-$R correctness estimate from $-1.7$ to $-4.0$ or $-2.3$, both with intervals excluding zero \\
\addlinespace[2pt]
Label choices made after inspecting results & Post hoc choices appear confirmatory & Seeds 4 and 5 for R and CT, the four-model pooled estimates, the small-cluster correction and the 24-transition analysis are labelled post hoc \\
\bottomrule
\end{tabularx}
\end{table}

\clearpage %
\bibliography{references}

\appendix
\aesbackmattersection{AI-use statement}
Generative AI tools assisted with software implementation, data analysis, literature discovery, and manuscript drafting and editing, but did not replace humans. The latter are responsible for the entire contents of this paper.

\section{Design details}
\label{app:design}
Table~\ref{tab:arms} lists what each arm receives, and Table~\ref{tab:design} sets out the three datasets used for the experiments.

\begin{table}[!htbp]
\centering\footnotesize
\caption{Experimental arms. Every arm except B0 receives the same R-arm feedback instructions in addition to its own report.}
\label{tab:arms}
\begin{tabular}{@{}ll@{}}\toprule
Arm & Message after a turn that changes code \\
\midrule
B0 & none \\
R & general feedback instructions only (``placebo'') \\
C & complexity table and general feedback instructions \\
T & coverage table and general feedback instructions \\
CT & both tables, shown one after the other, and general feedback instructions \\
\bottomrule\end{tabular}
\end{table}

\begin{table}[!htbp]
\centering\footnotesize
\caption{The three experiments. Each runs the five arms of Table~\ref{tab:arms} with the same feedback wording under DeepSeek V4.1 Flash (2026-09-10 release), GLM-5.3 Flash (2026-08-26 release), Claude Haiku 5.5 and GPT-6 Luna. A cell is one combination of task, arm, model and seed. Counts are summed over the four models.}
\label{tab:design}
\renewcommand{\arraystretch}{1.5}
\begin{tabular}{@{}>{\raggedright\arraybackslash}p{0.11\linewidth}>{\raggedright\arraybackslash}p{0.24\linewidth}>{\raggedright\arraybackslash}p{0.27\linewidth}>{\raggedright\arraybackslash}p{0.28\linewidth}@{}}\toprule
 & Commit0 & SlopCodeBench & SWE-EVO \\
\midrule
Task & reconstruct a Python library from its specification & build a program over 3--8 checkpoints, each a specification that extends the previous one & update a published project to its next release using that release's problem statement \\
Tasks & 25 libraries & 35 problems & 35 transitions in 6 projects \\
Seeds & 3; 5 for R and CT & 2 & 2 \\
Cells launched\,/ analyzed & 1,900\,/\,1,850 & 1,400\,/\,1,318 & 1,400\,/\,1,400 \\
Budget & 2M processed tokens per cell & 1M processed tokens per checkpoint & 2M processed tokens and 90 minutes per cell \\
Context & one conversation & fresh at each checkpoint; workspace, own tests and notes kept; every checkpoint runs even if the previous one failed & one conversation \\
Correctness & pass fraction on the withheld official suite, at the last graded snapshot at or below the cap & strict pass fraction on the final checkpoint's hidden tests, regression tests included & fix rate on the final tree (Section~\ref{sec:outcomes}) \\
Source & \def\UrlBreaks{\do\/}\url{https://github.com/commit-0/commit0} & \def\UrlBreaks{\do\/}\url{https://github.com/SprocketLab/slop-code-bench} & \def\UrlBreaks{\do\/}\url{https://github.com/SWE-EVO/SWE-EVO} \\
\bottomrule\end{tabular}

\end{table}

\paragraph{Models and conditions.} On Commit0 and SlopCodeBench, Haiku and Luna ran after Deep\-Seek and GLM, using newer container images %
and their developers' own endpoints. On SWE-EVO all four models ran in one campaign, each pinned to one provider. In addition to expected cross-model variability, this is why we estimate every effect within a model and never compare levels across models.

\paragraph{Analysis populations.} On Commit0, a graded workspace on which the hidden suite cannot import or complete scores zero on the whole fixed set of held-out tests. There are 278 such cells (106 GLM, 71 Luna, 57 Haiku, 44 DeepSeek). The primary analysis keeps these zeros; without them, the five-seed CT$-$R pool on correctness is $+0.1$ points [$-1.3$, $+1.5$]. A cell whose reference solution also fails to grade counts as missing. SlopCodeBench uses 35 of its 36 problems, because for one, \verb|datagate|, we could not successfully get the coverage step measure its generated files; it was left out before any cell ran. We have automatic reruns after an infrastructure failure. For Haiku and Luna, we graded 34 cells after a harness defect stopped them at the final checkpoint, without allowing additional tokens. Their inclusion was a post hoc decision. A test-list mismatch also excludes \verb|test_translator| from both models' analyses. All 1,400 SWE-EVO cells are graded. Before inspecting any T or CT result, we decided to exclude six \verb|dvc| tasks whose coverage step failed during test collection, leaving 29 transitions for contrasts involving T or CT. In four of them a test module needs a \verb|networkx| release that does not load under Python~3.9; in the other two it loads a storage plugin built for a newer \verb|fsspec| than the one installed. Either failure stops pytest before any test runs. Grading runs only the test files each task names, none of which fails to import, so every arm was graded normally. Coverage also failed on five of these 29 transitions. We retain the planned analysis and use the remaining 24 transitions from five projects for a post hoc sensitivity analysis. In that analysis, the pooled CT$-$R erosion estimate remains below zero ($-4.59$\textperthousand\ [$-5.27$, $-3.92$]), and the pooled fix-rate estimate falls to $+0.5$ points [$-2.0$, $+2.9$].

\paragraph{Estimands.} Tests within a task are not independent~\citep{miller2024errorbars}, so the unit of analysis is the task: a Commit0 library, a SlopCodeBench problem or a SWE-EVO transition. For each model, we first average seeds within each task and arm (between 2 and 5 seeds, depending on the task). We then average the paired differences, such as $\overline Y_{\ell,\mathrm{CT}}-\overline Y_{\ell,\mathrm R}$ for task $\ell$, giving equal weight to tasks with outcomes in both arms. We report a point estimate and paired $t$-test confidence interval for each model.
SWE-EVO's 35 transitions come from six projects; 21 belong to a single project. The specification calls for intervals clustered by project. We later chose a bias-reduced variance estimate (CR2) with Bell and McCaffrey's degrees of freedom~\citep{bell2002bias,pustejovsky2018small}, about two per model. The pooled estimate is an inverse-variance fixed-effect average of the four models. %
Its standard error assumes independent models, although they share tasks. Every pooled C$-$R and CT$-$R erosion estimate remains below zero when we resample whole tasks (whole projects on SWE-EVO) or fit the models jointly. $I^2$~\citep{higgins2002quantifying} estimates the proportion of variation among models attributable to heterogeneity.

\paragraph{SWE-EVO's secondary outcomes.} Resolution, defined as passing every official test, is a secondary outcome. Its pooled CT$-$R estimate is $+3.2$ points [$+0.7$, $+5.8$]. 
Pooled token use rises under CT by $+61$k per cell [$+34$k, $+89$k] and under T by $+43$k [$+28$k, $+59$k], with DeepSeek accounting for most of the T increase. These estimates include 25 cells whose token use is an upper-bound estimate. Excluding them gives CT$-$R of $+34$k [$-1$k, $+69$k] and T$-$R of $-8$k [$-15$k, $-1$k], leaving the effect on token use uncertain.

\paragraph{What measurement costs.} Overhead estimates use DeepSeek and GLM runs.%
Median message size ranges from 1.2\,kB for the reminder to 2.1\,kB for both reports. Injected text resulting from feedback accounts for a median 0.4 to 1.1\% of processed tokens per cell. Cyclomatic complexity scores, computed with \verb|lizard|~\citep{yin2026lizard}, are cached by file content, so a turn re-parses only the files it changed. Measuring coverage requires rerunning the agent's full test suite after each measured turn. On Commit0, this took a median 0.4 to 0.5\,s per turn, or 0.5 to 0.7\% of model time per cell. SlopCodeBench's suites took 9.6 to 16.9\,s per measured turn, with a median cost of 33 to 44\% of model time per cell. These figures exclude snapshotting and complexity measurement, which were not timed. Slow agent-written test suites can therefore make coverage measurement costly.

\paragraph{Integrity.} For the DeepSeek and GLM runs, an LLM (Opus 5.5) audited every trace for internet access, access to hidden tests or reference code, and grader tampering. On SlopCodeBench, the automated audit found no recorded access to hidden tests, reference code, grading outputs, another cell or the internet. Agents made many attempts, all blocked by operating-system isolation (Bubblewrap (\verb|bwrap|) inside Docker containers). On Commit0, the filesystem filter checked only path words. Agents routinely bypassed it by assembling paths from pieces, although the network proxy blocked every attempt to fetch upstream source. We excluded all 28 Commit0 cells in those runs that accessed hidden material or completed library code, including two identified by the audit. For Haiku and Luna, a scan for hidden-material access excluded 8 Commit0 cells (2 Haiku, 6 Luna). Auditing every SlopCodeBench trace for these models found no access to hidden material. On SWE-EVO, the hidden-test scan found no cell to exclude.
Both the scan and the audit can detect only activity recorded in the traces.

\paragraph{Mixed models.} The analysis in Section~\ref{sec:results} uses a binomial generalized linear mixed model of held-out test counts, fitted jointly across the four models after the initial analysis. It includes fixed effects for arm by model and for seed block or seed, random intercepts for task and task by model, and a cell intercept with model-specific variance.

\paragraph{What was fixed before the results.} Fixed in advance: the arms (including the wording of the feedback they provide), the primary contrasts (CT$-$R on correctness for Commit0 and SlopCodeBench; C$-$R, T$-$R and CT$-$R on the fix rate for SWE-EVO), %
and $E_{10}$, tokens and resolution as secondary outcomes. After inspecting the initial results, we added seeds 4 and 5 for R and CT on Commit0, pooled estimates across four models, the small-cluster correction on SWE-EVO, joint mixed models, the 24-transition sensitivity analysis and the 34 grades obtained without additional tokens. Each is labelled post hoc wherever it appears.

\subsection{Messages after a turn}
\label{app:messages}
After a turn that changes code and can be measured, arms R, C, T and CT receive one message; B0 receives none. R receives only the feedback instructions, headed \texttt{[Reminder, turn $k$]}. C, T and CT add measurement tables under \texttt{[Measurement, turn $k$]}. Each table lists changed functions above the limit in descending order (or \texttt{none}) and counts those at or below it. Every message ends with the same feedback instructions. The examples below are messages from Commit0 runs, with long tables truncated to their first rows.

\begin{trFeedbackBox}{R: reminder only}
\begin{verbatim}
[Reminder, turn 3]

Review the functions you changed this turn for readability, testability and
  maintainability.

For each function you changed:

1. Read the complete function and identify the distinct responsibilities it performs.
2. Simplify nested control flow using guard clauses or early returns where appropriate.
3. Extract coherent responsibilities into clearly named helper functions with explicit
  inputs and outputs.
4. Consolidate repeated branches or special cases using shared logic, lookup tables, or
  dispatch functions when appropriate.
5. Make sure the tests exercise the function's behavior, including its error paths and
  edge cases, and assert on results or effects.
6. Preserve the function's public behavior and interfaces, and run the relevant tests
  after any change, followed by the broader test suite when practical.

Do not create arbitrary one-line wrappers, scatter closely related logic across
  meaningless helpers, delete required behavior, or weaken tests. The goal is
  understandable and maintainable code.

Do not interrupt correct task completion for unnecessary refactoring. Correctness and
  passing the tests remain the first priority.
\end{verbatim}
\end{trFeedbackBox}

\begin{trFeedbackBox}{C: complexity table}
\begin{verbatim}
[Measurement, turn 1]

Cyclomatic complexity estimates how many independent execution paths a function contains.
  Higher complexity usually means the function is harder to understand, test, and modify
  safely. The target is 6 or lower for every function.

Functions above the limit, highest first:

  src/attr/_make.py:60  attrib
    cyclomatic complexity 18 (limit 6); new this turn

At or under the limit: 0 functions.

[the same feedback instructions as in the R message above]
\end{verbatim}
\end{trFeedbackBox}

\begin{trFeedbackBox}{T: uncovered changed lines}
\begin{verbatim}
[Measurement, turn 1]

Uncovered lines counts, for each function you changed this turn, the lines you added or
  edited that this repository's test suite did not execute. A line no test runs is a line
  whose behavior nothing checks. The target is 0 for every function you change.

Functions above the limit, highest first:

  jwt/utils.py:32  raw_to_der_signature
    uncovered lines 8 (limit 0): no test loads this file; new this turn
  jwt/utils.py:41  force_bytes
    uncovered lines 7 (limit 0): no test loads this file; new this turn
  jwt/utils.py:16  base64url_decode
    uncovered lines 6 (limit 0): no test loads this file; new this turn
  [...]

At or under the limit: 0 functions.

[the same feedback instructions as in the R message above]
\end{verbatim}
\end{trFeedbackBox}

\begin{trFeedbackBox}{CT: both tables}
\begin{verbatim}
[Measurement, turn 3]

Cyclomatic complexity estimates how many independent execution paths a function contains.
  Higher complexity usually means the function is harder to understand, test, and modify
  safely. The target is 6 or lower for every function.

Functions above the limit, highest first:

none

At or under the limit: 4 functions.

Uncovered lines counts, for each function you changed this turn, the lines you added or
  edited that this repository's test suite did not execute. A line no test runs is a line
  whose behavior nothing checks. The target is 0 for every function you change.

Functions above the limit, highest first:

  src/click/globals.py:8  get_current_context
    uncovered lines 6 (limit 0): no test loads this file; new this turn
  src/click/globals.py:40  resolve_color_default
    uncovered lines 6 (limit 0): no test loads this file; new this turn
  src/click/globals.py:30  push_context
    uncovered lines 3 (limit 0): no test loads this file; new this turn
  [...]

At or under the limit: 0 functions.

[the same feedback instructions as in the R message above]
\end{verbatim}
\end{trFeedbackBox}

\section{Every contrast}
\label{app:tables}
Tables~\ref{tab:app-erosion} to~\ref{tab:app-correctness-swe} report all four contrasts for erosion and correctness, separately by model and pooled post hoc. Table~\ref{tab:app-decomposition} reports the decomposition audit. In an illustrative manual review of the 20 largest reductions from DeepSeek and GLM runs, 17 divided the code into named, testable concepts. The response audit in Table~\ref{tab:app-response} tracks every edited function that exceeded the complexity limit or lacked full coverage, in every included cell on Commit0 and SlopCodeBench, whether or not a message flagged it. This response audit did not cover SWE-EVO.

\begin{figure}[!htbp]
\centering
\includegraphics[width=\linewidth]{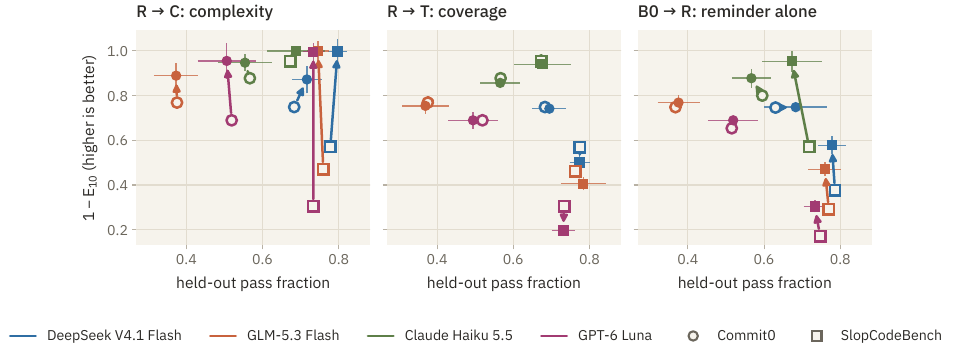}
\caption{Arrow diagrams for the other three contrasts, drawn as in Figure~\ref{fig:mitl-loop}(b): \textbf{(a)}~R$\to$C, complexity feedback; \textbf{(b)}~R$\to$T, coverage feedback; \textbf{(c)}~B0$\to$R, the reminder alone; four models, circles Commit0 and squares SlopCodeBench. Hollow markers denote the starting arm and filled markers the ending arm (Commit0: seeds 1--5 for R and 1--3 for C, T and B0; SlopCodeBench: complete-case pairing of problems within each model). Each arrow starts at the starting arm's mean for the paired tasks and shows the estimated paired change. Whiskers show the paired 95\% $t$ interval for that change, shifted to the arrowhead. They do not represent uncertainty in the ending arm's mean or a joint two-dimensional region. Each axis uses its own complete-case task set, so an arrow's pass fraction and $E_{10}$ may use different tasks. Absolute levels are not comparable across models.}
\label{fig:app-arrows-other}
\end{figure}

\begin{figure}[!htbp]
\centering
\includegraphics[width=\linewidth]{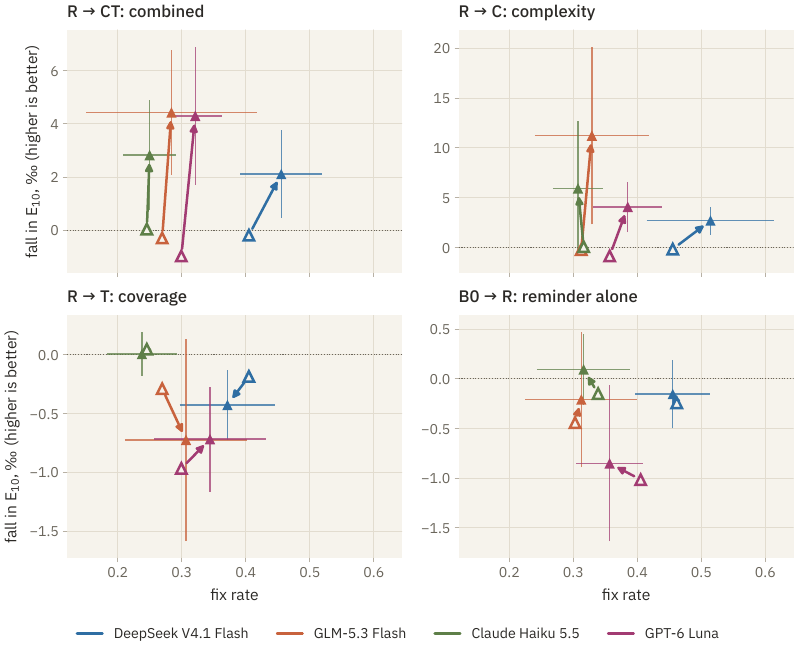}
\caption{SWE-EVO arrow diagrams for all four contrasts, drawn as in Figure~\ref{fig:app-arrows-other}, with the fix rate on the horizontal axis and the fall in $E_{10}$ from the starting code, in \textperthousand, on the vertical one; higher is better on both. Hollow and filled triangles denote the starting and ending arms, averaged over the paired transitions: 35 for R$\to$C and B0$\to$R, and 29 for R$\to$CT and R$\to$T. Arrows show the reported paired estimates. Whiskers show the paired 95\% $t$ intervals over transitions, shifted to the arrowhead. An interval containing zero therefore reaches the hollow triangle's coordinate on that axis. These plotted intervals treat transitions as independent. The main analysis instead clusters intervals by project (Section~\ref{sec:design}), giving wider intervals for most models and contrasts. Transition-level intervals exclude zero for seven of eight C$-$R and CT$-$R erosion estimates; project-clustered intervals exclude zero for four (Table~\ref{tab:lessons}). Vertical scales differ by panel.}
\label{fig:app-arrows-swe}
\end{figure}

Figure~\ref{fig:app-arrows-other} shows the other three contrasts on Commit0 and SlopCodeBench in the format of Figure~\ref{fig:mitl-loop}(b). Figure~\ref{fig:app-arrows-swe} shows all four contrasts on SWE-EVO.

\begin{table}[p]
\centering\footnotesize
\setlength{\tabcolsep}{3pt}
\caption{Change in $E_{10}$ at the final tree, per model, for all four contrasts. Lower is better. R gives the reminder-arm mean over its included tasks, with the task count in parentheses. Contrasts may use different paired task sets, so adding a contrast to R need not recover the other arm's mean. $n$ is the number of tasks paired for CT--R. Cells with no measured outcome are dropped. Commit0 pairs arm means within library, over seeds 1--5 for R and CT and seeds 1--3 for B0, C and T; SlopCodeBench pairs arm means within problem, over two seeds. $E_{10}$ was planned as a secondary outcome on all three benchmarks; correctness was the primary one. On SWE-EVO, contrasts involving T or CT use 29 of the 35 transitions. On six dvc transitions, a test module that fails to import stopped the project's suite before any test ran, so the coverage feedback that defines T and CT measured nothing. We excluded those six before reading any T or CT result (App.~\ref{app:design}); the other contrasts use all 35. Intervals are clustered by the six projects, as planned. The small-cluster correction was chosen later: CR2 with Bell--McCaffrey degrees of freedom~\citep{bell2002bias,pustejovsky2018small}, about two per model. The pooled estimate uses a $t$ interval with Satterthwaite degrees of freedom~\citep{satterthwaite1946approximate}. The pooled estimate is a post hoc fixed-effect inverse-variance average across four models. Its standard error assumes independent models despite their shared tasks. $I^2$ estimates the proportion of variation among the four estimates attributable to heterogeneity; high values indicate that the pooled estimate averages differing effects.}
\label{tab:app-erosion}
\begin{tabular}{lcccccc}\toprule
Model & R & C--R & CT--R & T--R & R--B0 & $n$ \\
\midrule
\multicolumn{7}{@{}l}{\textit{Commit0, 25 libraries}} \\
DeepSeek & \begin{tabular}[t]{@{}c@{}}0.251\\[-1pt]{\scriptsize (25)}\end{tabular} & \begin{tabular}[t]{@{}c@{}}\textminus{}0.123\\[-1pt]{\scriptsize [\textminus{}0.185, \textminus{}0.061]}\end{tabular} & \begin{tabular}[t]{@{}c@{}}\textminus{}0.070\\[-1pt]{\scriptsize [\textminus{}0.110, \textminus{}0.030]}\end{tabular} & \begin{tabular}[t]{@{}c@{}}+0.007\\[-1pt]{\scriptsize [\textminus{}0.009, +0.024]}\end{tabular} & \begin{tabular}[t]{@{}c@{}}\textminus{}0.002\\[-1pt]{\scriptsize [\textminus{}0.019, +0.015]}\end{tabular} & 25 \\
GLM & \begin{tabular}[t]{@{}c@{}}0.231\\[-1pt]{\scriptsize (24)}\end{tabular} & \begin{tabular}[t]{@{}c@{}}\textminus{}0.120\\[-1pt]{\scriptsize [\textminus{}0.177, \textminus{}0.064]}\end{tabular} & \begin{tabular}[t]{@{}c@{}}\textminus{}0.096\\[-1pt]{\scriptsize [\textminus{}0.145, \textminus{}0.046]}\end{tabular} & \begin{tabular}[t]{@{}c@{}}+0.015\\[-1pt]{\scriptsize [\textminus{}0.020, +0.049]}\end{tabular} & \begin{tabular}[t]{@{}c@{}}\textminus{}0.021\\[-1pt]{\scriptsize [\textminus{}0.056, +0.014]}\end{tabular} & 24 \\
Haiku & \begin{tabular}[t]{@{}c@{}}0.122\\[-1pt]{\scriptsize (25)}\end{tabular} & \begin{tabular}[t]{@{}c@{}}\textminus{}0.070\\[-1pt]{\scriptsize [\textminus{}0.108, \textminus{}0.031]}\end{tabular} & \begin{tabular}[t]{@{}c@{}}\textminus{}0.054\\[-1pt]{\scriptsize [\textminus{}0.082, \textminus{}0.026]}\end{tabular} & \begin{tabular}[t]{@{}c@{}}+0.021\\[-1pt]{\scriptsize [+0.002, +0.040]}\end{tabular} & \begin{tabular}[t]{@{}c@{}}\textminus{}0.078\\[-1pt]{\scriptsize [\textminus{}0.121, \textminus{}0.035]}\end{tabular} & 25 \\
Luna & \begin{tabular}[t]{@{}c@{}}0.311\\[-1pt]{\scriptsize (25)}\end{tabular} & \begin{tabular}[t]{@{}c@{}}\textminus{}0.266\\[-1pt]{\scriptsize [\textminus{}0.346, \textminus{}0.186]}\end{tabular} & \begin{tabular}[t]{@{}c@{}}\textminus{}0.261\\[-1pt]{\scriptsize [\textminus{}0.339, \textminus{}0.182]}\end{tabular} & \begin{tabular}[t]{@{}c@{}}0.000\\[-1pt]{\scriptsize [\textminus{}0.038, +0.038]}\end{tabular} & \begin{tabular}[t]{@{}c@{}}\textminus{}0.036\\[-1pt]{\scriptsize [\textminus{}0.056, \textminus{}0.015]}\end{tabular} & 25 \\
Pool (post hoc) & & \begin{tabular}[t]{@{}c@{}}\textminus{}0.112\\[-1pt]{\scriptsize [\textminus{}0.138, \textminus{}0.087]}\end{tabular} & \begin{tabular}[t]{@{}c@{}}\textminus{}0.078\\[-1pt]{\scriptsize [\textminus{}0.097, \textminus{}0.059]}\end{tabular} & \begin{tabular}[t]{@{}c@{}}+0.012\\[-1pt]{\scriptsize [+0.001, +0.023]}\end{tabular} & \begin{tabular}[t]{@{}c@{}}\textminus{}0.020\\[-1pt]{\scriptsize [\textminus{}0.032, \textminus{}0.009]}\end{tabular} & \\
\quad $I^2$ & & 86\% & 89\% & 0\% & 80\% & \\
\addlinespace[5pt] \multicolumn{7}{@{}l}{\textit{SlopCodeBench, 35 problems}} \\
DeepSeek & \begin{tabular}[t]{@{}c@{}}0.430\\[-1pt]{\scriptsize (35)}\end{tabular} & \begin{tabular}[t]{@{}c@{}}\textminus{}0.428\\[-1pt]{\scriptsize [\textminus{}0.483, \textminus{}0.372]}\end{tabular} & \begin{tabular}[t]{@{}c@{}}\textminus{}0.433\\[-1pt]{\scriptsize [\textminus{}0.487, \textminus{}0.379]}\end{tabular} & \begin{tabular}[t]{@{}c@{}}+0.067\\[-1pt]{\scriptsize [+0.030, +0.104]}\end{tabular} & \begin{tabular}[t]{@{}c@{}}\textminus{}0.202\\[-1pt]{\scriptsize [\textminus{}0.242, \textminus{}0.162]}\end{tabular} & 32 \\
GLM & \begin{tabular}[t]{@{}c@{}}0.530\\[-1pt]{\scriptsize (35)}\end{tabular} & \begin{tabular}[t]{@{}c@{}}\textminus{}0.529\\[-1pt]{\scriptsize [\textminus{}0.573, \textminus{}0.485]}\end{tabular} & \begin{tabular}[t]{@{}c@{}}\textminus{}0.504\\[-1pt]{\scriptsize [\textminus{}0.546, \textminus{}0.461]}\end{tabular} & \begin{tabular}[t]{@{}c@{}}+0.056\\[-1pt]{\scriptsize [+0.027, +0.085]}\end{tabular} & \begin{tabular}[t]{@{}c@{}}\textminus{}0.180\\[-1pt]{\scriptsize [\textminus{}0.210, \textminus{}0.149]}\end{tabular} & 34 \\
Haiku & \begin{tabular}[t]{@{}c@{}}0.046\\[-1pt]{\scriptsize (34)}\end{tabular} & \begin{tabular}[t]{@{}c@{}}\textminus{}0.045\\[-1pt]{\scriptsize [\textminus{}0.066, \textminus{}0.025]}\end{tabular} & \begin{tabular}[t]{@{}c@{}}\textminus{}0.045\\[-1pt]{\scriptsize [\textminus{}0.065, \textminus{}0.025]}\end{tabular} & \begin{tabular}[t]{@{}c@{}}+0.014\\[-1pt]{\scriptsize [\textminus{}0.008, +0.036]}\end{tabular} & \begin{tabular}[t]{@{}c@{}}\textminus{}0.383\\[-1pt]{\scriptsize [\textminus{}0.430, \textminus{}0.335]}\end{tabular} & 34 \\
Luna & \begin{tabular}[t]{@{}c@{}}0.696\\[-1pt]{\scriptsize (34)}\end{tabular} & \begin{tabular}[t]{@{}c@{}}\textminus{}0.692\\[-1pt]{\scriptsize [\textminus{}0.733, \textminus{}0.652]}\end{tabular} & \begin{tabular}[t]{@{}c@{}}\textminus{}0.687\\[-1pt]{\scriptsize [\textminus{}0.729, \textminus{}0.645]}\end{tabular} & \begin{tabular}[t]{@{}c@{}}+0.108\\[-1pt]{\scriptsize [+0.084, +0.133]}\end{tabular} & \begin{tabular}[t]{@{}c@{}}\textminus{}0.133\\[-1pt]{\scriptsize [\textminus{}0.160, \textminus{}0.106]}\end{tabular} & 34 \\
Pool (post hoc) & & \begin{tabular}[t]{@{}c@{}}\textminus{}0.246\\[-1pt]{\scriptsize [\textminus{}0.261, \textminus{}0.230]}\end{tabular} & \begin{tabular}[t]{@{}c@{}}\textminus{}0.236\\[-1pt]{\scriptsize [\textminus{}0.252, \textminus{}0.221]}\end{tabular} & \begin{tabular}[t]{@{}c@{}}+0.057\\[-1pt]{\scriptsize [+0.044, +0.070]}\end{tabular} & \begin{tabular}[t]{@{}c@{}}\textminus{}0.191\\[-1pt]{\scriptsize [\textminus{}0.207, \textminus{}0.174]}\end{tabular} & \\
\quad $I^2$ & & 100\% & 100\% & 91\% & 97\% & \\
\addlinespace[5pt] \multicolumn{7}{@{}l}{\textit{SWE-EVO, change in $E_{10}$ from the starting code, in \textperthousand}} \\
DeepSeek & \begin{tabular}[t]{@{}c@{}}+0.15\\[-1pt]{\scriptsize (35)}\end{tabular} & \begin{tabular}[t]{@{}c@{}}\textminus{}2.87\\[-1pt]{\scriptsize [\textminus{}5.22, \textminus{}0.51]}\end{tabular} & \begin{tabular}[t]{@{}c@{}}\textminus{}2.30\\[-1pt]{\scriptsize [\textminus{}4.96, +0.36]}\end{tabular} & \begin{tabular}[t]{@{}c@{}}+0.24\\[-1pt]{\scriptsize [\textminus{}0.13, +0.61]}\end{tabular} & \begin{tabular}[t]{@{}c@{}}\textminus{}0.09\\[-1pt]{\scriptsize [\textminus{}0.29, +0.11]}\end{tabular} & 29 \\
GLM & \begin{tabular}[t]{@{}c@{}}+0.21\\[-1pt]{\scriptsize (35)}\end{tabular} & \begin{tabular}[t]{@{}c@{}}\textminus{}11.46\\[-1pt]{\scriptsize [\textminus{}31.04, +8.12]}\end{tabular} & \begin{tabular}[t]{@{}c@{}}\textminus{}4.72\\[-1pt]{\scriptsize [\textminus{}7.02, \textminus{}2.43]}\end{tabular} & \begin{tabular}[t]{@{}c@{}}+0.44\\[-1pt]{\scriptsize [\textminus{}1.03, +1.90]}\end{tabular} & \begin{tabular}[t]{@{}c@{}}\textminus{}0.23\\[-1pt]{\scriptsize [\textminus{}0.71, +0.25]}\end{tabular} & 29 \\
Haiku & \begin{tabular}[t]{@{}c@{}}\textminus{}0.10\\[-1pt]{\scriptsize (35)}\end{tabular} & \begin{tabular}[t]{@{}c@{}}\textminus{}5.85\\[-1pt]{\scriptsize [\textminus{}14.75, +3.05]}\end{tabular} & \begin{tabular}[t]{@{}c@{}}\textminus{}2.78\\[-1pt]{\scriptsize [\textminus{}6.91, +1.34]}\end{tabular} & \begin{tabular}[t]{@{}c@{}}+0.04\\[-1pt]{\scriptsize [\textminus{}0.08, +0.16]}\end{tabular} & \begin{tabular}[t]{@{}c@{}}\textminus{}0.24\\[-1pt]{\scriptsize [\textminus{}0.63, +0.15]}\end{tabular} & 29 \\
Luna & \begin{tabular}[t]{@{}c@{}}+0.85\\[-1pt]{\scriptsize (35)}\end{tabular} & \begin{tabular}[t]{@{}c@{}}\textminus{}4.93\\[-1pt]{\scriptsize [\textminus{}8.46, \textminus{}1.40]}\end{tabular} & \begin{tabular}[t]{@{}c@{}}\textminus{}5.27\\[-1pt]{\scriptsize [\textminus{}8.76, \textminus{}1.79]}\end{tabular} & \begin{tabular}[t]{@{}c@{}}\textminus{}0.25\\[-1pt]{\scriptsize [\textminus{}1.26, +0.76]}\end{tabular} & \begin{tabular}[t]{@{}c@{}}\textminus{}0.16\\[-1pt]{\scriptsize [\textminus{}0.92, +0.59]}\end{tabular} & 29 \\
Pool (post hoc) & & \begin{tabular}[t]{@{}c@{}}\textminus{}3.68\\[-1pt]{\scriptsize [\textminus{}4.95, \textminus{}2.41]}\end{tabular} & \begin{tabular}[t]{@{}c@{}}\textminus{}3.85\\[-1pt]{\scriptsize [\textminus{}4.78, \textminus{}2.92]}\end{tabular} & \begin{tabular}[t]{@{}c@{}}+0.06\\[-1pt]{\scriptsize [\textminus{}0.04, +0.15]}\end{tabular} & \begin{tabular}[t]{@{}c@{}}\textminus{}0.14\\[-1pt]{\scriptsize [\textminus{}0.24, \textminus{}0.03]}\end{tabular} & \\
\quad $I^2$ & & 62\% & 67\% & 46\% & 0\% & \\
\bottomrule\end{tabular}
\end{table}

\begin{table}[p]
\centering\footnotesize
\setlength{\tabcolsep}{3pt}
\caption{Change in held-out pass fraction on Commit0 and SlopCodeBench, per model, in percentage points, for all four contrasts. Commit0 uses the withheld official suite at the 2M-token cap. SlopCodeBench uses the strict hidden-test pass fraction at the final checkpoint. R is expressed as a percentage. R gives the reminder-arm mean over its included tasks, with the task count in parentheses. Contrasts may use different paired task sets, so adding a contrast to R need not recover the other arm's mean. $n$ is the number of tasks paired for CT--R. Cells with no measured outcome are dropped. We added seeds 4 and 5 for R and CT after inspecting results from seeds 1--3. The other arms use seeds 1--3. The pooled estimate is a post hoc fixed-effect inverse-variance average across four models. Its standard error assumes independent models despite their shared tasks. $I^2$ estimates the proportion of variation among the four estimates attributable to heterogeneity; high values indicate that the pooled estimate averages differing effects.}
\label{tab:app-correctness}
\begin{tabular}{lcccccc}\toprule
Model & R & C--R & CT--R & T--R & R--B0 & $n$ \\
\midrule
\multicolumn{7}{@{}l}{\textit{Commit0, seeds 1--3 (planned)}} \\
DeepSeek & \begin{tabular}[t]{@{}c@{}}68.9\\[-1pt]{\scriptsize (25)}\end{tabular} & \begin{tabular}[t]{@{}c@{}}+2.7\\[-1pt]{\scriptsize [\textminus{}1.8, +7.3]}\end{tabular} & \begin{tabular}[t]{@{}c@{}}+4.4\\[-1pt]{\scriptsize [\textminus{}1.2, +10.0]}\end{tabular} & \begin{tabular}[t]{@{}c@{}}+0.6\\[-1pt]{\scriptsize [\textminus{}5.2, +6.3]}\end{tabular} & \begin{tabular}[t]{@{}c@{}}+5.9\\[-1pt]{\scriptsize [\textminus{}1.9, +13.7]}\end{tabular} & 25 \\
GLM & \begin{tabular}[t]{@{}c@{}}35.9\\[-1pt]{\scriptsize (25)}\end{tabular} & \begin{tabular}[t]{@{}c@{}}+1.4\\[-1pt]{\scriptsize [\textminus{}6.3, +9.1]}\end{tabular} & \begin{tabular}[t]{@{}c@{}}+4.7\\[-1pt]{\scriptsize [\textminus{}1.6, +11.0]}\end{tabular} & \begin{tabular}[t]{@{}c@{}}+1.0\\[-1pt]{\scriptsize [\textminus{}7.1, +9.0]}\end{tabular} & \begin{tabular}[t]{@{}c@{}}\textminus{}0.8\\[-1pt]{\scriptsize [\textminus{}8.2, +6.6]}\end{tabular} & 25 \\
Haiku & \begin{tabular}[t]{@{}c@{}}58.1\\[-1pt]{\scriptsize (25)}\end{tabular} & \begin{tabular}[t]{@{}c@{}}\textminus{}2.7\\[-1pt]{\scriptsize [\textminus{}9.9, +4.6]}\end{tabular} & \begin{tabular}[t]{@{}c@{}}\textminus{}2.3\\[-1pt]{\scriptsize [\textminus{}10.2, +5.6]}\end{tabular} & \begin{tabular}[t]{@{}c@{}}\textminus{}1.5\\[-1pt]{\scriptsize [\textminus{}7.0, +3.9]}\end{tabular} & \begin{tabular}[t]{@{}c@{}}\textminus{}1.5\\[-1pt]{\scriptsize [\textminus{}6.8, +3.9]}\end{tabular} & 25 \\
Luna & \begin{tabular}[t]{@{}c@{}}51.0\\[-1pt]{\scriptsize (25)}\end{tabular} & \begin{tabular}[t]{@{}c@{}}\textminus{}0.4\\[-1pt]{\scriptsize [\textminus{}8.8, +8.0]}\end{tabular} & \begin{tabular}[t]{@{}c@{}}+3.3\\[-1pt]{\scriptsize [\textminus{}4.8, +11.5]}\end{tabular} & \begin{tabular}[t]{@{}c@{}}\textminus{}1.6\\[-1pt]{\scriptsize [\textminus{}7.8, +4.6]}\end{tabular} & \begin{tabular}[t]{@{}c@{}}\textminus{}0.5\\[-1pt]{\scriptsize [\textminus{}6.5, +5.5]}\end{tabular} & 25 \\
Pool (post hoc) & & \begin{tabular}[t]{@{}c@{}}+1.0\\[-1pt]{\scriptsize [\textminus{}2.0, +4.0]}\end{tabular} & \begin{tabular}[t]{@{}c@{}}+3.1\\[-1pt]{\scriptsize [\textminus{}0.1, +6.3]}\end{tabular} & \begin{tabular}[t]{@{}c@{}}\textminus{}0.6\\[-1pt]{\scriptsize [\textminus{}3.5, +2.4]}\end{tabular} & \begin{tabular}[t]{@{}c@{}}+0.2\\[-1pt]{\scriptsize [\textminus{}2.9, +3.2]}\end{tabular} & \\
\quad $I^2$ & & 0\% & 0\% & 0\% & 0\% & \\
\addlinespace[5pt] \multicolumn{7}{@{}l}{\textit{Commit0, seeds 1--5 for R and CT (added after the first reading)}} \\
DeepSeek & \begin{tabular}[t]{@{}c@{}}68.3\\[-1pt]{\scriptsize (25)}\end{tabular} & \begin{tabular}[t]{@{}c@{}}+3.3\\[-1pt]{\scriptsize [\textminus{}0.5, +7.2]}\end{tabular} & \begin{tabular}[t]{@{}c@{}}+2.8\\[-1pt]{\scriptsize [\textminus{}1.3, +6.9]}\end{tabular} & \begin{tabular}[t]{@{}c@{}}+1.2\\[-1pt]{\scriptsize [\textminus{}3.3, +5.6]}\end{tabular} & \begin{tabular}[t]{@{}c@{}}+5.3\\[-1pt]{\scriptsize [\textminus{}3.0, +13.6]}\end{tabular} & 25 \\
GLM & \begin{tabular}[t]{@{}c@{}}37.5\\[-1pt]{\scriptsize (25)}\end{tabular} & \begin{tabular}[t]{@{}c@{}}\textminus{}0.2\\[-1pt]{\scriptsize [\textminus{}6.1, +5.6]}\end{tabular} & \begin{tabular}[t]{@{}c@{}}+2.6\\[-1pt]{\scriptsize [\textminus{}2.1, +7.3]}\end{tabular} & \begin{tabular}[t]{@{}c@{}}\textminus{}0.6\\[-1pt]{\scriptsize [\textminus{}6.7, +5.4]}\end{tabular} & \begin{tabular}[t]{@{}c@{}}+0.8\\[-1pt]{\scriptsize [\textminus{}4.7, +6.3]}\end{tabular} & 25 \\
Haiku & \begin{tabular}[t]{@{}c@{}}56.7\\[-1pt]{\scriptsize (25)}\end{tabular} & \begin{tabular}[t]{@{}c@{}}\textminus{}1.3\\[-1pt]{\scriptsize [\textminus{}8.3, +5.8]}\end{tabular} & \begin{tabular}[t]{@{}c@{}}\textminus{}0.5\\[-1pt]{\scriptsize [\textminus{}6.0, +5.0]}\end{tabular} & \begin{tabular}[t]{@{}c@{}}\textminus{}0.1\\[-1pt]{\scriptsize [\textminus{}5.4, +5.2]}\end{tabular} & \begin{tabular}[t]{@{}c@{}}\textminus{}2.9\\[-1pt]{\scriptsize [\textminus{}8.0, +2.3]}\end{tabular} & 25 \\
Luna & \begin{tabular}[t]{@{}c@{}}51.9\\[-1pt]{\scriptsize (25)}\end{tabular} & \begin{tabular}[t]{@{}c@{}}\textminus{}1.3\\[-1pt]{\scriptsize [\textminus{}9.0, +6.4]}\end{tabular} & \begin{tabular}[t]{@{}c@{}}+2.0\\[-1pt]{\scriptsize [\textminus{}3.4, +7.4]}\end{tabular} & \begin{tabular}[t]{@{}c@{}}\textminus{}2.5\\[-1pt]{\scriptsize [\textminus{}9.0, +4.0]}\end{tabular} & \begin{tabular}[t]{@{}c@{}}+0.4\\[-1pt]{\scriptsize [\textminus{}6.3, +7.0]}\end{tabular} & 25 \\
Pool (post hoc) & & \begin{tabular}[t]{@{}c@{}}+1.3\\[-1pt]{\scriptsize [\textminus{}1.3, +3.9]}\end{tabular} & \begin{tabular}[t]{@{}c@{}}+2.0\\[-1pt]{\scriptsize [\textminus{}0.3, +4.3]}\end{tabular} & \begin{tabular}[t]{@{}c@{}}\textminus{}0.2\\[-1pt]{\scriptsize [\textminus{}2.7, +2.4]}\end{tabular} & \begin{tabular}[t]{@{}c@{}}0.0\\[-1pt]{\scriptsize [\textminus{}2.9, +2.9]}\end{tabular} & \\
\quad $I^2$ & & 0\% & 0\% & 0\% & 6\% & \\
\addlinespace[5pt] \multicolumn{7}{@{}l}{\textit{SlopCodeBench, two seeds}} \\
DeepSeek & \begin{tabular}[t]{@{}c@{}}77.8\\[-1pt]{\scriptsize (35)}\end{tabular} & \begin{tabular}[t]{@{}c@{}}+1.9\\[-1pt]{\scriptsize [\textminus{}0.7, +4.5]}\end{tabular} & \begin{tabular}[t]{@{}c@{}}\textminus{}2.4\\[-1pt]{\scriptsize [\textminus{}6.1, +1.2]}\end{tabular} & \begin{tabular}[t]{@{}c@{}}\textminus{}0.1\\[-1pt]{\scriptsize [\textminus{}2.7, +2.5]}\end{tabular} & \begin{tabular}[t]{@{}c@{}}\textminus{}0.8\\[-1pt]{\scriptsize [\textminus{}4.5, +2.8]}\end{tabular} & 32 \\
GLM & \begin{tabular}[t]{@{}c@{}}76.0\\[-1pt]{\scriptsize (35)}\end{tabular} & \begin{tabular}[t]{@{}c@{}}\textminus{}1.4\\[-1pt]{\scriptsize [\textminus{}4.2, +1.4]}\end{tabular} & \begin{tabular}[t]{@{}c@{}}0.0\\[-1pt]{\scriptsize [\textminus{}3.1, +3.2]}\end{tabular} & \begin{tabular}[t]{@{}c@{}}+2.1\\[-1pt]{\scriptsize [\textminus{}3.8, +8.0]}\end{tabular} & \begin{tabular}[t]{@{}c@{}}\textminus{}1.1\\[-1pt]{\scriptsize [\textminus{}5.4, +3.3]}\end{tabular} & 34 \\
Haiku & \begin{tabular}[t]{@{}c@{}}67.3\\[-1pt]{\scriptsize (34)}\end{tabular} & \begin{tabular}[t]{@{}c@{}}+1.5\\[-1pt]{\scriptsize [\textminus{}6.1, +9.1]}\end{tabular} & \begin{tabular}[t]{@{}c@{}}\textminus{}5.8\\[-1pt]{\scriptsize [\textminus{}15.9, +4.3]}\end{tabular} & \begin{tabular}[t]{@{}c@{}}+0.4\\[-1pt]{\scriptsize [\textminus{}7.1, +7.9]}\end{tabular} & \begin{tabular}[t]{@{}c@{}}\textminus{}4.5\\[-1pt]{\scriptsize [\textminus{}12.3, +3.3]}\end{tabular} & 34 \\
Luna & \begin{tabular}[t]{@{}c@{}}73.3\\[-1pt]{\scriptsize (34)}\end{tabular} & \begin{tabular}[t]{@{}c@{}}0.0\\[-1pt]{\scriptsize [\textminus{}2.8, +2.8]}\end{tabular} & \begin{tabular}[t]{@{}c@{}}\textminus{}2.8\\[-1pt]{\scriptsize [\textminus{}6.3, +0.8]}\end{tabular} & \begin{tabular}[t]{@{}c@{}}\textminus{}0.2\\[-1pt]{\scriptsize [\textminus{}3.3, +2.9]}\end{tabular} & \begin{tabular}[t]{@{}c@{}}\textminus{}1.4\\[-1pt]{\scriptsize [\textminus{}4.3, +1.6]}\end{tabular} & 34 \\
Pool (post hoc) & & \begin{tabular}[t]{@{}c@{}}+0.3\\[-1pt]{\scriptsize [\textminus{}1.2, +1.8]}\end{tabular} & \begin{tabular}[t]{@{}c@{}}\textminus{}1.7\\[-1pt]{\scriptsize [\textminus{}3.6, +0.1]}\end{tabular} & \begin{tabular}[t]{@{}c@{}}+0.1\\[-1pt]{\scriptsize [\textminus{}1.6, +1.9]}\end{tabular} & \begin{tabular}[t]{@{}c@{}}\textminus{}1.4\\[-1pt]{\scriptsize [\textminus{}3.3, +0.5]}\end{tabular} & \\
\quad $I^2$ & & 10\% & 0\% & 0\% & 0\% & \\
\bottomrule\end{tabular}
\end{table}

\begin{table}[t]
\centering\footnotesize
\setlength{\tabcolsep}{3pt}
\caption{SWE-EVO correctness per model, in percentage points, for all four contrasts. Fix rate is the fraction of initially failing tests that pass on the final code. It is set to zero if the agent breaks a test that passes both before and after the reference change. Resolution means every official test passes. R is in percent. On SWE-EVO, the plan named three main comparisons, all on fix rate: C--R, T--R and CT--R. Each has its own 95\% interval, with no correction for multiple comparisons. Contrasts involving T or CT use 29 of the 35 transitions. On six dvc transitions, a test module that fails to import stopped the project's suite before any test ran, so the coverage feedback that defines T and CT measured nothing. We excluded those six before reading any T or CT result (App.~\ref{app:design}); the other contrasts use all 35. Intervals are clustered by the six projects, as planned. The small-cluster correction was chosen later: CR2 with Bell--McCaffrey degrees of freedom~\citep{bell2002bias,pustejovsky2018small}, about two per model. The pooled estimate uses a $t$ interval with Satterthwaite degrees of freedom~\citep{satterthwaite1946approximate}. The pooled estimate is a post hoc fixed-effect inverse-variance average across four models. Its standard error assumes independent models despite their shared tasks. $I^2$ estimates the proportion of variation among the four estimates attributable to heterogeneity; high values indicate that the pooled estimate averages differing effects.}
\label{tab:app-correctness-swe}
\begin{tabular}{lcccccc}\toprule
Model & R & C--R & CT--R & T--R & R--B0 & $n$ \\
\midrule
\multicolumn{7}{@{}l}{\textit{Fix rate (primary)}} \\
DeepSeek & \begin{tabular}[t]{@{}c@{}}45.5\\[-1pt]{\scriptsize (35)}\end{tabular} & \begin{tabular}[t]{@{}c@{}}+5.9\\[-1pt]{\scriptsize [\textminus{}7.6, +19.4]}\end{tabular} & \begin{tabular}[t]{@{}c@{}}+5.1\\[-1pt]{\scriptsize [\textminus{}2.3, +12.5]}\end{tabular} & \begin{tabular}[t]{@{}c@{}}\textminus{}3.4\\[-1pt]{\scriptsize [\textminus{}19.0, +12.3]}\end{tabular} & \begin{tabular}[t]{@{}c@{}}\textminus{}0.6\\[-1pt]{\scriptsize [\textminus{}12.4, +11.2]}\end{tabular} & 29 \\
GLM & \begin{tabular}[t]{@{}c@{}}31.2\\[-1pt]{\scriptsize (35)}\end{tabular} & \begin{tabular}[t]{@{}c@{}}+1.7\\[-1pt]{\scriptsize [\textminus{}14.3, +17.6]}\end{tabular} & \begin{tabular}[t]{@{}c@{}}+1.5\\[-1pt]{\scriptsize [\textminus{}31.2, +34.1]}\end{tabular} & \begin{tabular}[t]{@{}c@{}}+3.7\\[-1pt]{\scriptsize [\textminus{}16.6, +24.1]}\end{tabular} & \begin{tabular}[t]{@{}c@{}}+1.0\\[-1pt]{\scriptsize [\textminus{}11.7, +13.7]}\end{tabular} & 29 \\
Haiku & \begin{tabular}[t]{@{}c@{}}31.6\\[-1pt]{\scriptsize (35)}\end{tabular} & \begin{tabular}[t]{@{}c@{}}\textminus{}0.9\\[-1pt]{\scriptsize [\textminus{}9.1, +7.3]}\end{tabular} & \begin{tabular}[t]{@{}c@{}}+0.4\\[-1pt]{\scriptsize [\textminus{}8.1, +9.0]}\end{tabular} & \begin{tabular}[t]{@{}c@{}}\textminus{}0.8\\[-1pt]{\scriptsize [\textminus{}11.4, +9.9]}\end{tabular} & \begin{tabular}[t]{@{}c@{}}\textminus{}2.3\\[-1pt]{\scriptsize [\textminus{}8.6, +4.1]}\end{tabular} & 29 \\
Luna & \begin{tabular}[t]{@{}c@{}}35.7\\[-1pt]{\scriptsize (35)}\end{tabular} & \begin{tabular}[t]{@{}c@{}}+2.8\\[-1pt]{\scriptsize [\textminus{}18.8, +24.5]}\end{tabular} & \begin{tabular}[t]{@{}c@{}}+2.2\\[-1pt]{\scriptsize [\textminus{}5.4, +9.7]}\end{tabular} & \begin{tabular}[t]{@{}c@{}}+4.5\\[-1pt]{\scriptsize [\textminus{}20.0, +28.9]}\end{tabular} & \begin{tabular}[t]{@{}c@{}}\textminus{}4.9\\[-1pt]{\scriptsize [\textminus{}25.6, +15.9]}\end{tabular} & 29 \\
Pool (post hoc) & & \begin{tabular}[t]{@{}c@{}}+1.2\\[-1pt]{\scriptsize [\textminus{}2.6, +4.9]}\end{tabular} & \begin{tabular}[t]{@{}c@{}}+2.7\\[-1pt]{\scriptsize [\textminus{}0.2, +5.6]}\end{tabular} & \begin{tabular}[t]{@{}c@{}}\textminus{}0.2\\[-1pt]{\scriptsize [\textminus{}5.3, +4.8]}\end{tabular} & \begin{tabular}[t]{@{}c@{}}\textminus{}1.6\\[-1pt]{\scriptsize [\textminus{}4.8, +1.5]}\end{tabular} & \\
\quad $I^2$ & & 9\% & 0\% & 0\% & 0\% & \\
\addlinespace[5pt] \multicolumn{7}{@{}l}{\textit{Resolution (secondary)}} \\
DeepSeek & \begin{tabular}[t]{@{}c@{}}31.4\\[-1pt]{\scriptsize (35)}\end{tabular} & \begin{tabular}[t]{@{}c@{}}+2.9\\[-1pt]{\scriptsize [\textminus{}15.6, +21.3]}\end{tabular} & \begin{tabular}[t]{@{}c@{}}+5.2\\[-1pt]{\scriptsize [\textminus{}0.9, +11.2]}\end{tabular} & \begin{tabular}[t]{@{}c@{}}\textminus{}3.4\\[-1pt]{\scriptsize [\textminus{}18.6, +11.7]}\end{tabular} & \begin{tabular}[t]{@{}c@{}}\textminus{}1.4\\[-1pt]{\scriptsize [\textminus{}10.9, +8.0]}\end{tabular} & 29 \\
GLM & \begin{tabular}[t]{@{}c@{}}18.6\\[-1pt]{\scriptsize (35)}\end{tabular} & \begin{tabular}[t]{@{}c@{}}+5.7\\[-1pt]{\scriptsize [\textminus{}7.2, +18.7]}\end{tabular} & \begin{tabular}[t]{@{}c@{}}+3.4\\[-1pt]{\scriptsize [\textminus{}17.0, +23.9]}\end{tabular} & \begin{tabular}[t]{@{}c@{}}+5.2\\[-1pt]{\scriptsize [\textminus{}12.1, +22.4]}\end{tabular} & \begin{tabular}[t]{@{}c@{}}0.0\\[-1pt]{\scriptsize [\textminus{}17.0, +17.0]}\end{tabular} & 29 \\
Haiku & \begin{tabular}[t]{@{}c@{}}20.0\\[-1pt]{\scriptsize (35)}\end{tabular} & \begin{tabular}[t]{@{}c@{}}\textminus{}1.4\\[-1pt]{\scriptsize [\textminus{}5.4, +2.5]}\end{tabular} & \begin{tabular}[t]{@{}c@{}}+1.7\\[-1pt]{\scriptsize [\textminus{}5.8, +9.3]}\end{tabular} & \begin{tabular}[t]{@{}c@{}}\textminus{}1.7\\[-1pt]{\scriptsize [\textminus{}12.0, +8.6]}\end{tabular} & \begin{tabular}[t]{@{}c@{}}\textminus{}4.3\\[-1pt]{\scriptsize [\textminus{}10.3, +1.7]}\end{tabular} & 29 \\
Luna & \begin{tabular}[t]{@{}c@{}}22.9\\[-1pt]{\scriptsize (35)}\end{tabular} & \begin{tabular}[t]{@{}c@{}}+1.4\\[-1pt]{\scriptsize [\textminus{}17.7, +20.5]}\end{tabular} & \begin{tabular}[t]{@{}c@{}}+1.7\\[-1pt]{\scriptsize [\textminus{}5.8, +9.3]}\end{tabular} & \begin{tabular}[t]{@{}c@{}}\textminus{}1.7\\[-1pt]{\scriptsize [\textminus{}17.1, +13.7]}\end{tabular} & \begin{tabular}[t]{@{}c@{}}\textminus{}2.9\\[-1pt]{\scriptsize [\textminus{}19.0, +13.2]}\end{tabular} & 29 \\
Pool (post hoc) & & \begin{tabular}[t]{@{}c@{}}\textminus{}0.6\\[-1pt]{\scriptsize [\textminus{}3.5, +2.3]}\end{tabular} & \begin{tabular}[t]{@{}c@{}}+3.2\\[-1pt]{\scriptsize [+0.7, +5.8]}\end{tabular} & \begin{tabular}[t]{@{}c@{}}\textminus{}1.0\\[-1pt]{\scriptsize [\textminus{}5.4, +3.4]}\end{tabular} & \begin{tabular}[t]{@{}c@{}}\textminus{}3.2\\[-1pt]{\scriptsize [\textminus{}6.1, \textminus{}0.2]}\end{tabular} & \\
\quad $I^2$ & & 47\% & 0\% & 0\% & 0\% & \\
\bottomrule\end{tabular}
\end{table}

\begin{table}[t]
\centering\footnotesize
\setlength{\tabcolsep}{3pt}
\caption{Agent responses to feedback, using included cells and Commit0 seeds 1--3. For complexity, we report the proportions of over-limit functions edited again and under the limit at the end of the run. C and CT count functions flagged in the report; R counts over-limit functions edited by the agent. For coverage, all arms use the same rule: among functions edited while coverage was below 1, we report the proportion whose coverage later increased. The proportions pool function-turn observations, so a function counts once for each turn on which it was flagged or edited. No uncertainty intervals are reported.}
\label{tab:app-response}
\begin{tabular}{llcccccc}\toprule
 & & \multicolumn{3}{c}{complexity: edited / under limit} & \multicolumn{3}{c}{coverage rose} \\
 Benchmark & Model & R & C & CT & R & T & CT \\
\midrule
Commit0 & DeepSeek & 20\% / 1\% & 54\% / 46\% & 36\% / 23\% & 64\% & 57\% & 57\% \\
Commit0 & GLM & 30\% / 3\% & 60\% / 58\% & 46\% / 35\% & 40\% & 42\% & 34\% \\
Commit0 & Haiku & 31\% / 9\% & 83\% / 86\% & 67\% / 66\% & 47\% & 51\% & 49\% \\
Commit0 & Luna & 49\% / 3\% & 88\% / 94\% & 85\% / 90\% & 59\% & 86\% & 81\% \\
\addlinespace[5pt] SCB & DeepSeek & 38\% / 5\% & 94\% / 94\% & 84\% / 82\% & 17\% & 95\% & 92\% \\
SCB & GLM & 41\% / 4\% & 90\% / 94\% & 82\% / 85\% & 10\% & 38\% & 63\% \\
SCB & Haiku & 44\% / 18\% & 87\% / 95\% & 81\% / 92\% & 49\% & 85\% & 83\% \\
SCB & Luna & 60\% / 4\% & 90\% / 97\% & 87\% / 95\% & 26\% & 84\% & 75\% \\
\bottomrule\end{tabular}
\end{table}

\begin{table}[t]
\centering\footnotesize
\setlength{\tabcolsep}{3pt}
\caption{What lies behind the drop in $E_{10}$. A branch is a point where the code can take one of two paths, such as an \texttt{if}, a loop, an \texttt{and} or an \texttt{or}; a function's cyclomatic complexity is its number of branches plus one. Splitting a complex function into helpers lowers $E_{10}$ without removing a single branch, so we check how much of that drop survives undoing it. Each pair is one task and seed with both arms' final code on record (seeds 1--3 on Commit0). \emph{As measured} is the change in $E_{10}$ on these pairs; it differs slightly from Table~\ref{tab:app-erosion}, which averages seeds within each task first. \emph{After inlining} repeats the measurement as if every function called from only one place had been written inside its caller. We do this in both arms, including helpers written under R or by the libraries' original authors, so it is a sensitivity check rather than an estimate of what splitting causes. The last three columns describe the whole program. \emph{Branches} is the percent change in the number of branches, which splitting does not remove; \emph{lines} is the percent change in lines of code; and \emph{per 1k lines} is the change in branches per thousand lines of code. \emph{Branches} and \emph{per 1k lines} can have opposite signs because the code's length changes too: a few more branches spread over many more lines means fewer branches per line. The branch count also grows with how much of the library an agent builds, and a few pairs where the C or CT run built much more than the R run make it rise in some rows. Intervals are 95\% bootstrap intervals that resample whole tasks; the two percent changes have none.}
\label{tab:app-decomposition}
\begin{tabular}{@{}llcccrrc@{}}\toprule
 & & & \multicolumn{2}{c}{change in $E_{10}$} & \multicolumn{3}{c}{change in branches and lines} \\
Model & & pairs & as measured & after inlining & branches & lines & per 1k lines \\
\midrule
\multicolumn{8}{@{}l}{\textit{Commit0, seeds 1--3}} \\
DeepSeek & C--R & 62 & \textminus{}0.134 {\scriptsize[\textminus{}0.200, \textminus{}0.073]} & \textminus{}0.023 {\scriptsize[\textminus{}0.043, \textminus{}0.004]} & +1\% & +3\% & \textminus{}7.8 {\scriptsize[\textminus{}14.0, \textminus{}1.9]} \\
 & CT--R & 61 & \textminus{}0.066 {\scriptsize[\textminus{}0.103, \textminus{}0.035]} & \textminus{}0.016 {\scriptsize[\textminus{}0.037, +0.004]} & +1\% & +2\% & \textminus{}5.6 {\scriptsize[\textminus{}10.3, \textminus{}0.8]} \\
GLM & C--R & 56 & \textminus{}0.124 {\scriptsize[\textminus{}0.190, \textminus{}0.060]} & \textminus{}0.029 {\scriptsize[\textminus{}0.085, +0.020]} & 0\% & +4\% & \textminus{}4.6 {\scriptsize[\textminus{}13.2, +5.6]} \\
 & CT--R & 55 & \textminus{}0.096 {\scriptsize[\textminus{}0.166, \textminus{}0.025]} & \textminus{}0.003 {\scriptsize[\textminus{}0.047, +0.051]} & \textminus{}4\% & +1\% & \textminus{}5.3 {\scriptsize[\textminus{}17.3, +7.9]} \\
Haiku & C--R & 69 & \textminus{}0.065 {\scriptsize[\textminus{}0.102, \textminus{}0.033]} & \textminus{}0.011 {\scriptsize[\textminus{}0.038, +0.016]} & +4\% & +6\% & \textminus{}6.0 {\scriptsize[\textminus{}16.8, +4.7]} \\
 & CT--R & 69 & \textminus{}0.054 {\scriptsize[\textminus{}0.087, \textminus{}0.026]} & \textminus{}0.025 {\scriptsize[\textminus{}0.059, +0.005]} & \textminus{}6\% & +3\% & \textminus{}16.5 {\scriptsize[\textminus{}31.2, \textminus{}5.1]} \\
Luna & C--R & 74 & \textminus{}0.268 {\scriptsize[\textminus{}0.337, \textminus{}0.198]} & \textminus{}0.075 {\scriptsize[\textminus{}0.099, \textminus{}0.050]} & \textminus{}16\% & \textminus{}4\% & \textminus{}22.9 {\scriptsize[\textminus{}30.0, \textminus{}16.6]} \\
 & CT--R & 74 & \textminus{}0.265 {\scriptsize[\textminus{}0.333, \textminus{}0.197]} & \textminus{}0.087 {\scriptsize[\textminus{}0.114, \textminus{}0.061]} & \textminus{}16\% & \textminus{}4\% & \textminus{}21.6 {\scriptsize[\textminus{}29.1, \textminus{}14.8]} \\
\addlinespace[5pt] \multicolumn{8}{@{}l}{\textit{SlopCodeBench, two seeds}} \\
DeepSeek & C--R & 64 & \textminus{}0.424 {\scriptsize[\textminus{}0.476, \textminus{}0.369]} & \textminus{}0.044 {\scriptsize[\textminus{}0.066, \textminus{}0.022]} & \textminus{}16\% & \textminus{}7\% & \textminus{}18.2 {\scriptsize[\textminus{}28.8, \textminus{}7.9]} \\
 & CT--R & 58 & \textminus{}0.422 {\scriptsize[\textminus{}0.468, \textminus{}0.373]} & \textminus{}0.062 {\scriptsize[\textminus{}0.085, \textminus{}0.039]} & \textminus{}21\% & \textminus{}10\% & \textminus{}27.5 {\scriptsize[\textminus{}39.5, \textminus{}16.4]} \\
GLM & C--R & 62 & \textminus{}0.527 {\scriptsize[\textminus{}0.566, \textminus{}0.486]} & \textminus{}0.076 {\scriptsize[\textminus{}0.098, \textminus{}0.058]} & \textminus{}16\% & \textminus{}3\% & \textminus{}35.4 {\scriptsize[\textminus{}44.6, \textminus{}26.0]} \\
 & CT--R & 56 & \textminus{}0.515 {\scriptsize[\textminus{}0.551, \textminus{}0.475]} & \textminus{}0.095 {\scriptsize[\textminus{}0.120, \textminus{}0.071]} & \textminus{}21\% & \textminus{}6\% & \textminus{}37.5 {\scriptsize[\textminus{}46.5, \textminus{}28.1]} \\
Haiku & C--R & 64 & \textminus{}0.043 {\scriptsize[\textminus{}0.063, \textminus{}0.026]} & \textminus{}0.029 {\scriptsize[\textminus{}0.049, \textminus{}0.010]} & \textminus{}4\% & \textminus{}7\% & +5.5 {\scriptsize[\textminus{}1.1, +12.0]} \\
 & CT--R & 65 & \textminus{}0.046 {\scriptsize[\textminus{}0.066, \textminus{}0.027]} & \textminus{}0.053 {\scriptsize[\textminus{}0.088, \textminus{}0.024]} & \textminus{}18\% & \textminus{}19\% & +3.0 {\scriptsize[\textminus{}3.7, +9.3]} \\
Luna & C--R & 68 & \textminus{}0.692 {\scriptsize[\textminus{}0.730, \textminus{}0.654]} & \textminus{}0.063 {\scriptsize[\textminus{}0.078, \textminus{}0.049]} & \textminus{}14\% & \textminus{}4\% & \textminus{}30.0 {\scriptsize[\textminus{}40.3, \textminus{}20.4]} \\
 & CT--R & 68 & \textminus{}0.687 {\scriptsize[\textminus{}0.726, \textminus{}0.646]} & \textminus{}0.054 {\scriptsize[\textminus{}0.067, \textminus{}0.041]} & \textminus{}17\% & \textminus{}9\% & \textminus{}23.8 {\scriptsize[\textminus{}33.9, \textminus{}14.7]} \\
\bottomrule\end{tabular}
\end{table}

\clearpage
\section{Erosion along the run}
\label{app:trajectories}
\begin{figure}[!htbp]
\centering
\includegraphics[width=\linewidth]{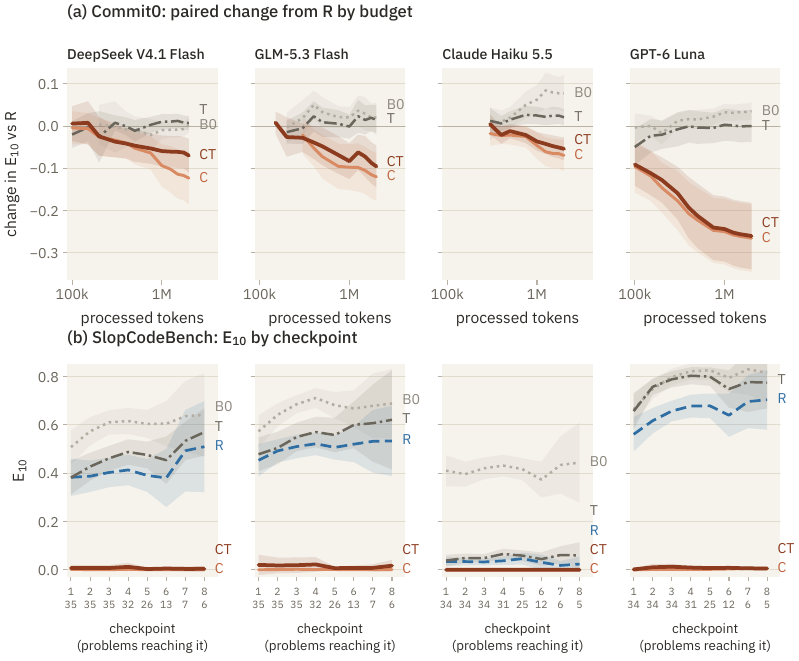}
\caption{$E_{10}$ for every arm and model. \textbf{(a)}~Commit0: paired change in $E_{10}$ from R at each processed-token budget, with libraries as pairs and pointwise 95\% $t$ bands. Points require at least 15 paired libraries. R and CT use seeds 1--5; the other arms use seeds 1--3. \textbf{(b)}~SlopCodeBench: $E_{10}$ by arm at each checkpoint, averaged first within problems and then across problems, with a pointwise 95\% $t$ band. We use included cells from the main runs, with two seeds. Points require at least 5 problems at the checkpoint. The second row of tick labels gives the number reaching it under any arm for that model.}
\label{fig:app-trajectories-models}
\end{figure}

Figure~\ref{fig:trajectories} in the main text follows CT$-$R through each run. Figure~\ref{fig:app-trajectories-models} shows every arm. On Commit0, erosion under C and CT falls further below R as tokens are used. T and B0 remain close to R, although B0 ends above R for Haiku and Luna, as does T for Haiku. On SlopCodeBench, C and CT keep $E_{10}$ at or below 0.022 at every checkpoint for all four models.

\begin{figure}[!htbp]
\centering
\includegraphics[width=\linewidth]{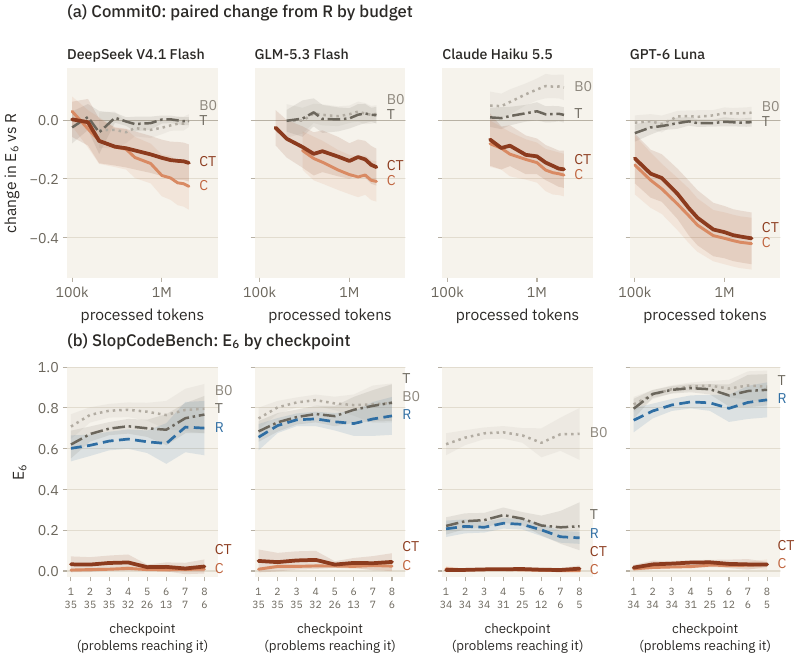}
\caption{$E_6$, the share of complexity mass in functions above the report's limit of 6, for every arm and model, drawn as Figure~\ref{fig:app-trajectories-models} draws $E_{10}$. \textbf{(a)}~Commit0: paired change in $E_6$ from R at each processed-token budget, with libraries as pairs and pointwise 95\% $t$ bands. Points require at least 15 paired libraries. R and CT use seeds 1--5; the other arms use seeds 1--3. \textbf{(b)}~SlopCodeBench: $E_6$ at each checkpoint by arm, built as in Figure~\ref{fig:app-trajectories-models}(b).}
\label{fig:app-e6}
\end{figure}

\paragraph{A check on the stated limit.} The complexity report asks the agent to bring every changed function to complexity 6 or below, and $E_6$, the share of complexity mass above that limit, measures the request directly. We use $E_6$ as a manipulation check and retain $E_{10}$ as the outcome. Under C and CT, $E_6$ ends below R on both benchmarks for every model, and on SlopCodeBench it stays close to zero at every checkpoint, while T and B0 stay with R or above it (Figure~\ref{fig:app-e6}). The reminder does not state the limit. An arm that states it without measuring code would test whether the target alone changes behavior (Section~\ref{sec:conclusion}).

\end{document}